\documentclass[lettersize,journal]{IEEEtran}
\usepackage{amsmath,amsfonts}
\usepackage{algorithmic}
\usepackage{algorithm}
\usepackage{array}
\usepackage[caption=false,font=normalsize,labelfont=sf,textfont=sf]{subfig}
\usepackage{textcomp}
\usepackage{stfloats}
\usepackage{url}
\usepackage{verbatim}
\usepackage{graphicx}
\usepackage{wrapfig}
\usepackage{cite}
\usepackage{graphicx}
\usepackage{amssymb}
\usepackage{booktabs}
\usepackage[utf8]{inputenc} % allow utf-8 input
\usepackage[T1]{fontenc}    % use 8-bit T1 fonts
\usepackage{url}            % simple URL typesetting
\usepackage{nicefrac}       % compact symbols for 1/2, etc.
\usepackage{microtype}      % microtypography
\usepackage{lipsum}
\usepackage{fancyhdr}       % header
\usepackage{graphicx}       % graphics
\usepackage{xcolor}         % colors
\usepackage{bbm}
\usepackage{mathtools}
\usepackage{pifont}
\newcommand{\xmark}{\text{\ding{55}}}

\usepackage[hidelinks,colorlinks=true,linkcolor=blue,citecolor=blue]{hyperref}

\begin{document}

\title{Visualizing and Understanding Contrastive Learning

\thanks{This research received funding from the FWO (Grants G014718N, G0A4720N and 1SB5721N) and from imec through the AAA project Trustworthy AI Methods (TAIM).

F. Sammani, B. Joukovsky and N. Deligiannis are with the Department of Electronics and Informatics, Vrije Universiteit Brussel, Pleinlaan 2, B-1050 Brussels, Belgium and also with imec, Kapeldreef 75, B-3001 Leuven, Belgium.
}
}

\author{Fawaz Sammani,~\IEEEmembership{Student Member,~IEEE,} Boris Joukovsky,~\IEEEmembership{Student Member,~IEEE,} and~Nikos~Deligiannis,~\IEEEmembership{Member,~IEEE}}



\maketitle

\begin{abstract}
Contrastive learning has revolutionized the field of computer vision, learning rich representations from unlabeled data, which generalize well to diverse vision tasks. Consequently, it has become increasingly important to explain these approaches and understand their inner workings mechanisms. Given that contrastive models are trained with interdependent and interacting inputs and aim to learn invariance through data augmentation, the existing methods for explaining single-image systems (\textit{e.g.,} image classification models) are inadequate as they fail to account for these factors and typically assume independent inputs. Additionally, there is a lack of evaluation metrics designed to assess pairs of explanations, and no analytical studies have been conducted to investigate the effectiveness of different techniques used to explaining contrastive learning. In this work, we design visual explanation methods that contribute towards understanding similarity learning tasks from pairs of images. We further adapt existing metrics, used to evaluate visual explanations of image classification systems, to suit pairs of explanations and evaluate our proposed methods with these metrics. Finally, we present a thorough analysis of visual explainability methods for contrastive learning, establish their correlation with downstream tasks and demonstrate the potential of our approaches to investigate their merits and drawbacks.
\end{abstract}


\section{Introduction}
\label{sec:intro}
\IEEEPARstart{C}{\lowercase{ontrastive learning}} is a technique used to train machine learning models to recognize similarities between two inputs. One input is designated as a positive sample and the other as a negative sample \cite{Koch2015SiameseNN}. The aim of contrastive learning is to maximize the similarity between positive inputs while minimizing it between negative ones. Recently, contrastive learning has revolutionized the field of model pretraining \cite{chen2020big}, \cite{Radford2021LearningTV},\cite{He2020MomentumCF},\cite{Caron2021EmergingPI}, which involves learning representations from large amounts of unlabeled data. Pretrained models have demonstrated notable benefits over supervised learning by effectively adapting to specific downstream tasks through fine-tuning, while also learning more general, non-biased features. In recent years, contrastive learning has become increasingly reliant on data augmentations to generate two distinct views of the same image, which are used as positive samples. This characteristic makes contrastive learning a fully unsupervised task, as it does not require positive or negative mining \cite{Lin2017FocalLF},\cite{Shrivastava2016TrainingRO}. 

\begin{figure}
    \centering
    \includegraphics[width=0.5\textwidth]{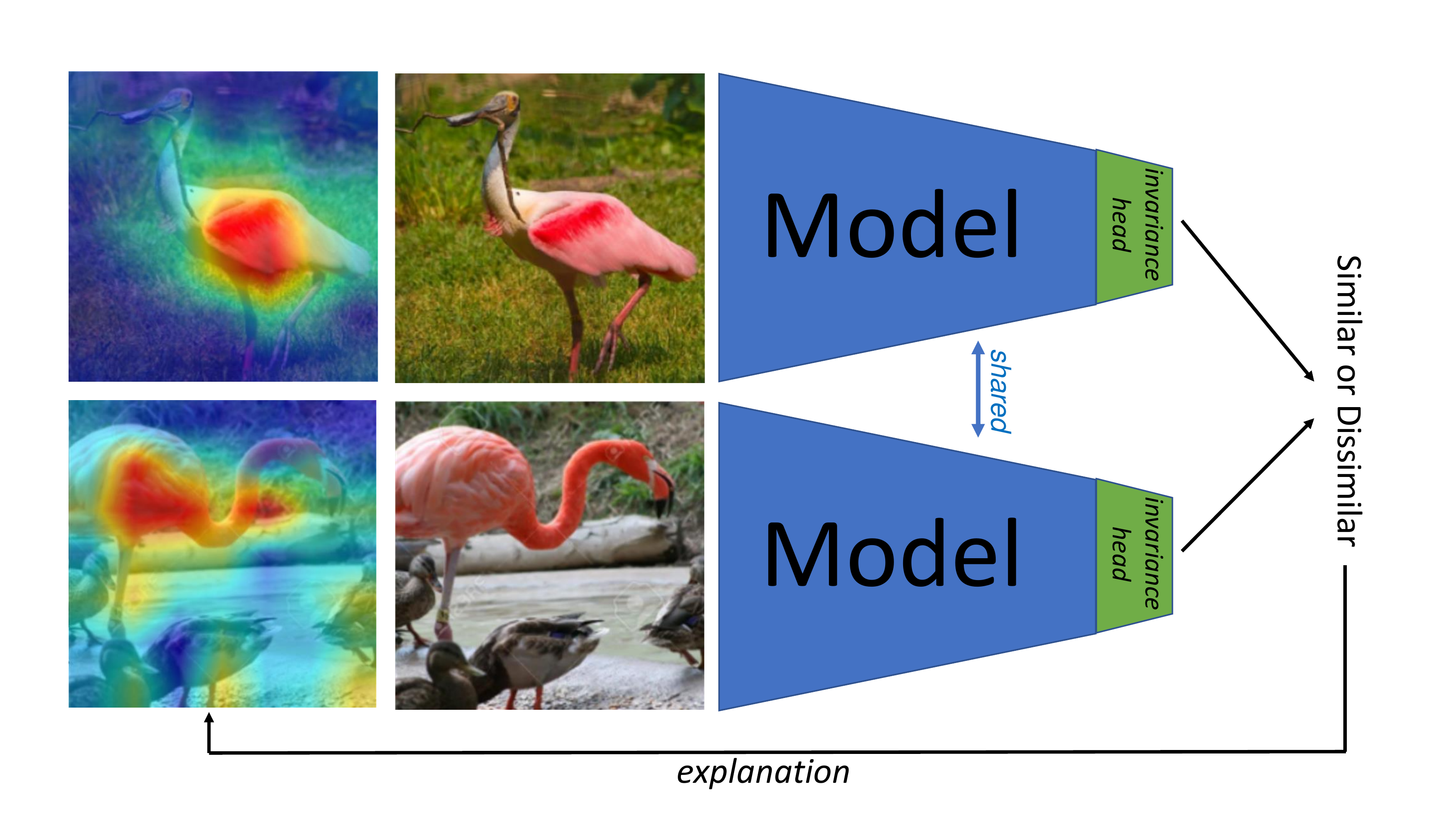}
    \caption{Contrastive models compare pairs of inputs and determine whether they are similar or not. Our work involves the development of methods to explain how these models make decisions and how they learn to recognize similarities.}
    \vspace{-0.5cm}
    \label{demoTIP}
\end{figure}

Explainable Artificial Intelligence (XAI) involves building techniques to understand the underlying mechanisms behind the decision-making process of a machine learning model. It is crucial for building trust and accountability in AI systems as it enables users to understand and improve their models \cite{Bargal2021GuidedZZ} and identify any data biases present \cite{Bach2016AnalyzingCF}, \cite{Lapuschkin2019UnmaskingCH}. This ensures fairness and equability \cite{Goodman2017EuropeanUR}. Local post-hoc attribution methods attribute what features are responsible for a specific prediction, and have gained significant popularity in recent years. They are roughly divided into three categories, mostly aimed towards explaining image classification models: perturbation-based \cite{Zeiler2014VisualizingAU}, \cite{Ribeiro2016WhySI}, activation-based \cite{Zhou2016LearningDF,Selvaraju2019GradCAMVE, Bach2015OnPE} and gradient-based methods \cite{Simonyan2014DeepIC, Smilkov2017SmoothGradRN,Sundararajan2017AxiomaticAF}. 

While XAI has made significant progress in explaining image classification models, its application to explaining contrastive learning models remains scarce. BiLRP \cite{Eberle2022BuildingAI} explains similarity models involving a pair of images using Layer-wise Relevance Propagation (LRP) \cite{Bach2015OnPE}, designing new rules which re-distribute the similarity score through the layers back to the input. However, it requires knowledge about which LRP-rule to be used at which layer, and new rules have to be derived for different models including LSTMs \cite{Arras2017ExplainingRN}, Batch Normalization \cite{Ioffe2015BatchNA, Guillemot2020BreakingBN}, Multi-Head Attention \cite{Vaswani2017AttentionIA, Voita2019AnalyzingMS}, and Transformers \cite{Chefer2021TransformerIB}. Another work, Similarity Decomposition \cite{Stylianou2019VisualizingDS} provides an explanation which reflects the similarity of features, by computing the dot product between each spatial location of one image and the average-pooled vector of the other image. However, this method is restricted to models which contrast the average-pooled features directly. This limitation hinders the explanation map to provide any information about the invariance encoded in the model. As the recent self-supervised contrastive models \cite{chen2020simple, chen2020big, Chen2021ExploringSS, Zbontar2021BarlowTS} learn similarity through invariance, this component cannot be overlooked. The projection head connecting the average-pooled features and the final contrasted features is responsible for encoding this invariance \cite{Chen2021AnES, Chen2021ExploringSS, chen2020simple}. Our proposed methods address this gap by considering features in both the similarity and invariance space. Contrastive learning can also be used to learn a new model that is tuned to be explainable for a given explanation method \cite{Pillai2022ConsistentEB}.

Given the increasing reliance on contrastive learning models, its importance cannot be overlooked. In this regard, we discuss several reasons for why it is crucial to explain contrastive models. First, this enables us to determine whether the contrastive learning task is effective and whether the model is able to learn meaningful concepts between the two images. Second, this allows us to gain a better understanding of the process of data augmentation (\textit{e.g.,} which augment better helps the model). Third, many models are based on learning from two inputs \textit{without} being finetuned. Image Steganography \cite{Baluja2017HidingII}, face recognition with Siamese networks \cite{Koch2015SiameseNN}, anomaly detection in videos \cite{Sultani2018RealWorldAD}, \cite{Kpkl2021DriverAD}, CrossVIT \cite{Chen2021CrossViTCM} and MultiView Transformer \cite{Yan2022MultiviewTF} are popular examples. Our techniques are directly applicable to these models as well. Fourth, since the pretrained model is used to initialize weights in the fine-tuning process, any data bias introduced in the pretrained model will also transfer to the fine-tuned model. Lastly, pretrained contrastive models before finetuning are usually used as feature extractors (\textit{e.g.,} CLIP \cite{Radford2021LearningTV}) for different computer vision applications \cite{Shen2022HowMC}, \cite{ Sammani2022NLXGPTAM}. All these reasons motivate us to visualize and understand contrastive models. 

Existing XAI methods designed for explaining models that learn from a single image (\textit{e.g.,} image classification models) are not well-suited to explain contrastive models. Unlike single-image systems, contrastive models processes information differently as they learn from shared, dependent, and entangled inputs that interact with each other. As a result, simply employing existing XAI techniques, which typically assume independent inputs, would lead to inaccurate explanations. Moreover, recent contrastive models \cite{chen2020big,Radford2021LearningTV,He2020MomentumCF,Caron2021EmergingPI} aim towards learning \textit{invariance} from data augmentations. That is, they should be invariant to input transformations.  Consequently, invariance should also be considered as a crucial aspect in the design of XAI methods for contrastive learning. Our work thus aims at developing techniques to explain the underlying mechanisms of contrastive learning, including understanding how they learn similarity and the rationale behind the decisions they make. Likewise, most evaluation metrics in the literature of explainability \cite{Petsiuk2018RISERI, Chattopadhyay2018GradCAMGG, Yeh2019HowSA, Wang2020ScoreCAMSV} are designed for single image systems. In the case of contrastive learning, the model takes a pair of images as input and outputs a similarity score. As a result, two explanation maps are obtained rather than a single one. The need for metrics to evaluate pairs of explanations is currently not fulfilled and therefore, such metrics are necessary. Hence, we additionally adapt existing metrics used to evaluate explanations of single-image models to a \textit{simultaneous} and \textit{conditional} setting, which accommodates the evaluation of pairs of explanations. Finally, there exists no analytical study on XAI techniques for contrastive learning. As a result, it is unclear which specific method or family of methods is optimal and most effective for specific purposes and use-cases. Our study is designed to be all-encompassing. It involves an examination of contrastive models and an overview of the effectiveness of diverse methods and their connection to downstream tasks. 

In summary, the contributions of this work are as follows:

\begin{itemize}
  \item We propose post-hoc visual explanation techniques for explaining contrastive vision models that learn from a pair of images. 
  \item We extend existing evaluation metrics designed for supervised learning tasks (especially, image classification) to suit pair of explanations and evaluate the proposed methods with these metrics. 
  \item We offer a comprehensive study and analysis of XAI techniques for contrastive learning. This includes a thorough exploration of contrastive models, a summary of the performance of various methods, their relationship to downstream tasks as well as some explorations and insights. This equips users with a comprehensive guide to understanding contrastive models.
\end{itemize}

The remainder of the paper is as follows: Section \ref{sec:related} reviews background information and Section \ref{sec:method} presents our explainability techniques. Section \ref{sec:eval} elaborates on the designed metrics to quantitatively evaluate our explanations. In Section \ref{sec:exp}, we perform both quantitative and qualitative analyses, present a summary of the results, and explore applications for our methods. Additionally, we examine the relationship between explainability and downstream task performance. Finally, we offer further insights and explorations in Section \ref{sec:findings}.

\section{Background on related XAI techniques}
\label{sec:related}

In this Section, we discuss earlier works in the literature of XAI for image classification systems, which we will re-visit in our work. 

\subsection{Saliency-based Methods} Saliency maps~\cite{Simonyan2014DeepIC} are methods for visualizing the importance of individual pixels of an image $I$ to any given neuron~$c$. A saliency map is defined as: $I \circ \frac{\partial c}{\partial I}$, where $\circ$ denotes the Hadamard product. A high gradient for a pixel in $I$ signifies that alterations of that pixel would increase the value of $c$. $c$ is usually defined as the neuron representing an output class. Saliency maps have been found to be noisy due to shattering gradients and saturation, resulting in the development of new techniques. Smooth-Grad \cite{Smilkov2017SmoothGradRN} removes noise from saliency maps by adding random Gaussian noise to the input image for several steps, producing local samples around the sample of interest and eliminating local variations in non-smoothly varying partial derivatives. Integrated Gradients \cite{Sundararajan2017AxiomaticAF} addresses the issue of gradient saturation which arrises when models are trained until saturation. This is achieved by desaturating the input image. Guided Backpropagation \cite{Springenberg2015StrivingFS} removes noise from saliency maps by backpropagating only the positive ReLU gradients. As a result, pixels that negatively correlate with the output score are not shown in the saliency map. 

\subsection{Occlusion Analysis} Occlusion analysis \cite{Zeiler2014VisualizingAU} is a fundamental and straightforward approach used to understand the significance of different regions in an image $I$ on a neuron $c$. The method involves the use of a black or gray mask to obscure specific regions of the image, and the resultant drop in $c$ relative to the baseline is directly measured and recorded. The reduction in score provides evidence that the obscured region is vital for prediction. The mask is applied using a sliding window approach to cover all areas of the image. This perturbation-based technique does not require access to model structure, weights, or gradients, having an advantage over other methods \cite{Bach2015OnPE,Kwon2021InverseBasedAT,Simonyan2014DeepIC, Selvaraju2019GradCAMVE}. 

\subsection{Class Activation Maps (CAM)} CAM \cite{Zhou2016LearningDF}, later generalized to Grad-CAM \cite{Selvaraju2019GradCAMVE}, is a visualization technique used to examine the activations of a specific convolutional layer in a neural network. Specifically, Grad-CAM considers an activation map $A\in \mathbb{R}^{K\times w\times h}$ of depth $K$ and spatial dimensions $w\times h$, and defines the importance map as $\sum_{k} [\alpha_k]_{+} A^{k}$. The notation $[\alpha_k]_{+}$ means only the positive values of $\alpha_k$ are retained. The coefficients $\alpha_k = \frac{1}{K} \sum_{k=1}^K \frac{\partial c}{\partial A^k}$ correspond to the spatially-averaged gradients of the model's output class $c$ with respect to (w.r.t.) $A^k$. Several variants of CAM have been proposed, primarily focused on developing new methods for weighting the activation maps, including Grad-CAM++ \cite{Chattopadhyay2018GradCAMGG}, Score-CAM \cite{Wang2020ScoreCAMSV}, and Relevance-CAM \cite{Lee2021RelevanceCAMYM}. However, the underlying principle remains the same. More recent extensions of CAM such as HighResCAM~\cite{Draelos2020UseHI} and LayerCAM~\cite{Jiang2021LayerCAMEH} proposed to directly weight the activation map $A^k$ point-wise with the non-averaged gradients, showing that this approach leads to better coverage of the salient objects in the importance map for earlier layers. 

\section{Proposed Methods}
\label{sec:method}
This section describes our proposed techniques for explaining contrastive vision models. We note that the methods are \textit{architecture-independent}, that is, they are applicable to both CNNs and Transformers. Our code is publicly available\footnote{https://github.com/annonymoussubmission/explain-cl}. Moreover, we provide a web-demo\footnote{https://huggingface.co/spaces/AnnonSubmission/xai-cl} to facilitate users in effectively visualizing and understanding different contrastive learning models.

\subsection{Saliency-based Methods}
\subsubsection{Baseline} We start by outlining a straightforward extension of Input $\times$ Grad \cite{Simonyan2014DeepIC} and Smooth-Grad \cite{Smilkov2017SmoothGradRN} to two-image models, which can serve as our baselines. A simple approach to extend these methods to two-image models would be to apply them separately to each image. In the case of Input $\times$ Grad, the produced explanations maps are: $I_{1} \circ \frac{\partial s}{\partial A_1}, I_{2} \circ \frac{\partial s}{\partial A_2} \in \mathbb{R}^{w \times h}$. However, given that contrastive models utilize different and random data augmentations to generate two views of the same image, this simple extension ignores how the change in augmentation of each pixel impacts the output score. We now describe our proposed method, coined~\textit{Average Transforms}.

\subsubsection{Averaged Transforms} 
Since numerous contrastive models are trained to be invariant to input transformations, it is essential to incorporate the invariance component into the the explainability algorithm. However, before doing so, we aim to validate the contrastive models' learning objective qualitatively. In the appendix, we present an experiment that demonstrates how invariance progresses for both contrastive models and image classification models. The experiment shows that as the layer depth increases, the image saturation and colors de-emphasize and decrease. This indicates that the latent representations learned by contrastive models are more invariant to input transformations than those of image classification models.

The objective of Averaged Transforms is to examine the impact of data augmentations and invariance on the similarity result. During the training process of contrastive models, data augmentation is applied to input samples to create diverse positive and negative pairs. A crucial aspect of data augmentation is the random level of strength applied to each augment, which affects the magnitude of the transformation. As an example, the widely used "color jitter" augment applies random brightness, contrast, saturation, and hue levels to distort image color. In the case of "Gaussian blur", random values within a range of standard deviations (strengths) are selected and used to create Gaussian kernels which blur the image. Consequently, all different strengths must be considered in order to fully understand the effect of a specific data augmentation technique on the model, since different strength values can potentially lead to different interpretations of the prediction.

The purpose of this experiment is to determine the effects that a particular data transformation $T$ has on two images. We start by decomposing a transformation $T$ into $[t^{1}, t^{2}, \dots, t^{Z}]$, where $t^{z}$ represent a specific strength of the transformation $T$ and $z$ gradually varies from the lowest to the highest strength possible that does not fully distort the image and cause a distribution shift. Let $I_{1}$ and $I_{2}$ be two input images, and let $t^{z}(I_{2})$ denote the $z$th transformation of $I_{2}$. We then compute the similarity score between $I_{1}$ and $t^{z}(I_{2})$ for all values of $z$ from $1$ to $Z$: $S_z=\operatorname{sim}\left(I_1, t^z\left(I_2\right)\right), \quad \forall z=1, \ldots, Z$, where $\mathrm{sim}$ is the cosine similarity function. We can then compute the gradient of the similarity score with respect to each of the input images $I_{1}$ and $I_{2}$: $\nabla_{I_1} S_z, \quad \nabla_{I_2} S_z, \quad \forall z=1, \ldots, Z$. Finally, we average the gradients across different strengths for each image to obtain a final saliency map: $\mathcal{S}_{I_1}=I_{1} \circ \frac{1}{Z} \sum_{z=1}^Z \nabla_{I_1} S_z, \quad \mathcal{S}_{I_2}= I_{2} \circ \frac{1}{Z} \sum_{z=1}^Z \nabla_{I_2} S_z$. $\mathcal{S}_{I_1}$ and $\mathcal{S}_{I_2}$ represent the saliency maps for $I_{1}$ and $I_{2}$, respectively.

Practically, we can implement the aforementioned process by applying $t^{Z}$ to $I_{2}$ directly, such that $I_{Z} = t^{Z}(I_{2})$, and then gradually transform $I_{1}$ to $I_{2}$ by applying $o^{z}=\rho I_{1}+(1-\rho) I_{Z}$, where $\rho$ is the step size that gradually decreases from 1 to 0 by a factor of 0.1. We may (optionally) choose to remove noise from the saliency map by using noise removal techniques such as Smooth-Grad \cite{Smilkov2017SmoothGradRN} or Guided Backpropagation \cite{Springenberg2015StrivingFS}. In Section \ref{sec:exp}, we experiment and ablate all possible options of our method (w/ and w/o Smooth-Grad, w/ and w/o Guided Backpropagation). Finally, we may also blur the pixel-wise saliency map output with a Gaussian Kernel for visually appealing results (not performed when evaluating the methods).

\subsection{Perturbation-based Methods}
\label{sec:perturbation}

\subsubsection{Baseline} We start by  describing a straightforward extension to two-image models coined \textit{Conditional Occlusion}: given the input images $I_1$ and $I_2$ and their similarity score~$s$, we successively zero-out parts of $I_1$ according to a square sliding-window, while $I_2$ remains unchanged. Let $I_1^{\operatorname{p}}$ denote the perturbed sample. For each window location, we record the output of the constrastive model fed with $(I_1^{\operatorname{p}}, I_2)$: a larger drop means that the occluded region is more important for deeming these two images as similar, conditioned on $I_2$ being fully observed. We  plot an importance map for $I_1$ by subtracting the scores of the perturbed images from the baseline $s$. The process can be carried out for the second image by swapping $I_1$ and $I_2$. For all quantitative and qualitative results, we use a perturbation mask size of 64 $\times$ 64 and a stride of 8. We now describe the problems of the baseline method and introduce our proposed method, coined \textit{Pairwise Occlusion}. 

\begin{figure}
    \centering
    \includegraphics[width=0.45\textwidth]{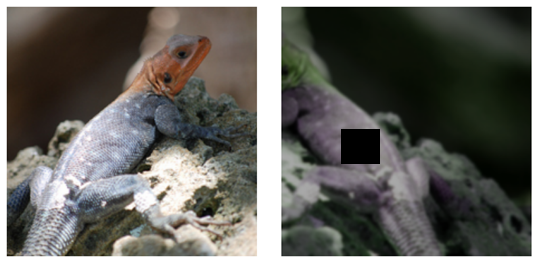}
    \caption{An example of computing conditional occlusion of the second image conditioned on the first image. The contextual information surrounding the dropped region (indicated by a black box) in the second image is sufficient to produce a high similarity prediction with the fully-observed first image.}
    \label{problem_condOcc}
\end{figure}

\begin{figure}
    \centering
    \includegraphics[width=0.5\textwidth]{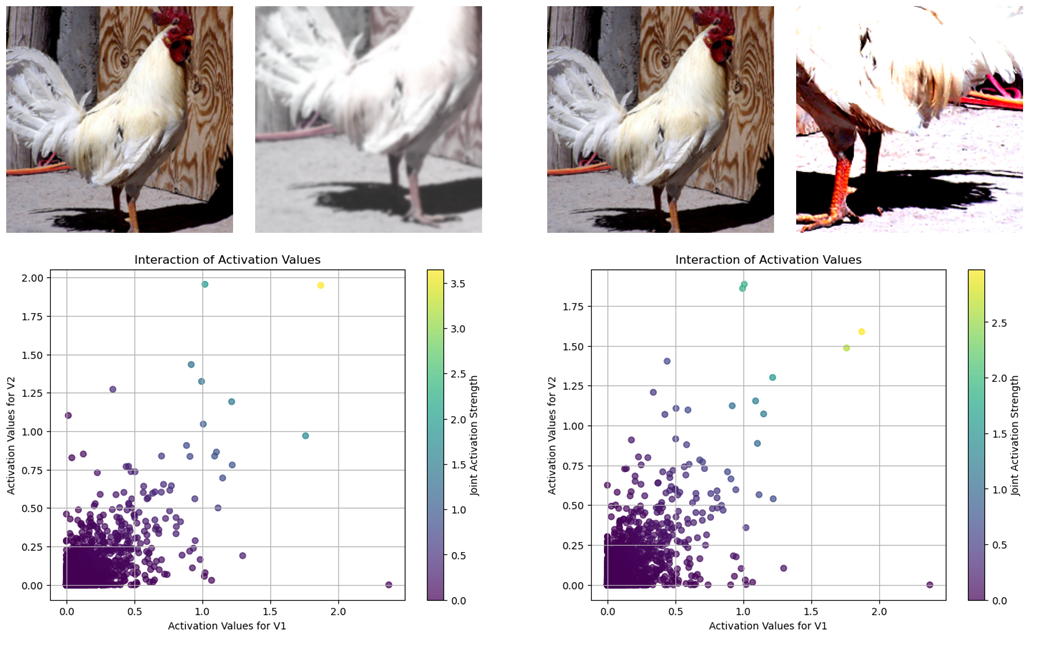}
    \caption{A 2D plot representing the interaction of activation vectors for augmented image pairs. The $x$-axis represents feature values from the first image's activation vector and the $y$-axis represents those of the second image. The color bar indicates the magnitude of their interaction (their element-wise multiplication).}
    \label{activation_interactions_plots}
\end{figure}

\subsubsection{Pairwise Occlusion} A problem with Conditional Occlusion is that the similarity score may not drop when perturbing a region because the context around the dropped region being present in both images may be sufficient for a high similarity prediction. This problem is illustrated in Fig. \ref{problem_condOcc}. Ideally, we would like a region to be considered significant only if its prediction cannot be inferred from its context. We propose an approach which considers simultaneous perturbations of the two images while mitigating two issues that may arise with this approach: (1) testing all the combinations of regions between the two images would result in a quadratic increase of required forward passes, and (2) less-significant regions might be given too much importance. We alleviate the first risk by randomly choosing different occlusions sizes, locations and aspect ratios for each image, which asymptotically approximates all possible combinations of perturbations. The second risk may occur when a drop in similarity is caused by removing either of the perturbed regions. As an example, consider two pictures of a dog on neutral backgrounds, the first is perturbed with an occlusion of the dog's face, and the second contains an occlusion in the background. In this case, the similarity drop is mainly caused by the occlusion of the dog's face, and therefore must be given more importance. To alleviate this issue, we perform the following: after obtaining the similarity scores of all perturbed pairs, we perform a softmax operation on the scores to obtain a distribution of importances among all different simultaneous perturbations. Then, for each pair, we assign the softmax score to the perturbed region of the image whose high-level features have the lowest $\ell_2$ norm, indicating that this region likely contained highly activated features. For all quantitative and qualitative results, we sampled $100$ different perturbation masks with different occlusion sizes, locations and aspect ratios. The scale of the mask ranges from $10\% - 30 \%$ of the image. 

\subsection{Activation Visualization Methods}

\subsubsection{Baseline} We start by describing a straightforward extension of Grad-CAM \cite{Selvaraju2019GradCAMVE} to two-image models, which serves as our baseline. We consider two activation maps $A_1, A_2\in \mathbb{R}^{K\times w \times h}$ obtained respectively by feeding the first and second input images to the contrastive model. Also, let $s$ denote the predicted similarity score. Next, we compute the gradient maps $\frac{\partial s}{\partial A_1}, \frac{\partial s}{\partial A_2} \in \mathbb{R}^{K\times w \times h}$. By this approach, the importance maps $E_1$ and $E_2$ are obtained separately by weighing the activation maps with their rectified gradients $W^k_{1,2} = \frac{\partial s}{\partial A^k_{1,2}}$. Eq.~\eqref{eq:baselinegradcam} thus shows the expressions of $E_1$ and $E_2$, and $\odot$ denotes the Hadamard product. We refer to this model as Baseline-Grad-CAM in the further experiments of Section~\ref{sec:exp}.
\begin{equation}\label{eq:baselinegradcam}
    E_{1,2} = \textstyle \sum_k [W_{1,2}^k]_{+} \odot A_{1,2} ^k.
\end{equation}

Notably, Baseline-Grad-CAM does not incorporate any joint term in the formulation, which is crucial for reflecting the similarity score derived from two input images. Baseline-Grad-CAM disentangles this joint process and explains each image independently of the other. As it treats each image separately, it misses out on capturing the interdependence between images learned by the contrastive model. 

\subsubsection{Interaction-CAM} We now describe our proposed method. Since the similarity score is computed by considering two input images, the interaction of their corresponding activations and gradients should be reflected by the produced explainability maps. Unlike existing methods that rely solely on the activations and gradients of individual images, our approach considers the joint activations and gradients of both images to produce explainability maps that highlight the common features and how they influence each other. Specifically, Interaction-CAM further augments the baseline method by introducing a joint activation term $J$ and a gradient interaction term $G$ to capture common activations and gradients between $A_1$ and $A_2$.

The joint activation $J = [j^1, \dots, j^K]$ is defined by the element-wise product of two $K$-dimensional vectors $J_1 \odot J_2$, where $j_{1,2}^k$ results from a spatial reduction operation of the corresponding feature map $A_{1,2}^k$. Therefore, each element of $J$ indicates whether both images simultaneously activate similar features. We provide a visual example in Fig.~\ref{activation_interactions_plots}. When a feature value from the activation vector is high for the first image but low for the second image (as shown by the data points in the bottom-right quadrant of both plots), the resulting interaction value is notably low (indicated by the purple hue). Conversely, the inverse scenario holds true, as shown by the data points in the top-left quadrant of the plot. This allows us to visualize important features which are jointly active in both images, and suppress features that are active in one image but low in the other. Later in Section~\ref{sec:exp}, we experiment and ablate three different reduction operations: average pooling, max pooling and self-attention with the query set as the average-pool of $A^{k}_{1,2}$, and the keys and values as $A^{k}_{1,2}$.

We define the global gradient interaction matrix $G\in\mathbb{R}^{K\times K}$ (a cross-correlation matrix of the gradients) by computing inner products of gradients across the spatial dimension, that is, $G^{kl} = \operatorname{vec}(\frac{\partial s}{\partial A_1^k})^T\operatorname{vec}(\frac{\partial s}{\partial A_2^l})$. The inner product thus encodes the similarity of the channel gradients a given spatial locations of the first and second image. It provides a measure to which extent the two gradient vectors are in the same direction. Then, for each feature map and for each input image, we retrieve a unique coefficient $g_{1,2}^k$ via a reduction operator on $G$ against the \textit{other} image: $g_{1}^{k} = \max_{l'}(G^{kl'})$ and $g_{2}^{k} = \max_{k'}(G^{k'l})$. Note that the maximum value defines a high similarity between the gradients of all pixels from both images at a specific dimension. In simple terms, the gradient interaction term encodes the interaction dynamics of gradients, specifically how the channel gradients of the two images affected each other.  

The two final importance maps are obtained by aggregating all feature maps, weighted by the associated joint activation and gradient interaction terms, as shown in Eq.~\eqref{eq:interactioncam}:
\begin{equation}\label{eq:interactioncam}
    E_{1,2} =\textstyle \sum_{k}[j^k\cdot g_{1,2}^k]_{+}\cdot A_{1,2}^{k}.
\end{equation}

\begin{table*}[]
\centering
\caption{Evaluation scores on perturbation, activation and saliency-based methods for SimCLRv2 (1$\times$) and SimCLRv2 (2$\times$). SI: Simultaneous Insertion, SD: Simultaneous Deletion, CI: Conditional Insertion, CD: Conditional Deletion, SAD: Simultaneous Average Drop, CAD: Conditional Average Drop. AR stands for the type we use for Activation Reduction and GI indicates whether or not we use Gradient Interaction. We ablate all possible combinations.}
\scalebox{0.9}{
\begin{tabular}{c|c|c|c|c|c|c|c|c|c|c|c|c|c|c}
\hline
\multicolumn{3}{c|}{} & \multicolumn{6}{c|}{SimCLRv2 (1$\times$)} & \multicolumn{6}{c}{SimCLRv2 (2$\times$)} \\
\hline
& AR & GI & \textbf{SI $\uparrow$} & \textbf{SD $\downarrow$} & \textbf{CI $\uparrow$} & \textbf{CD $\downarrow$} & \textbf{SAD $\downarrow$} & \textbf{CAD $\downarrow$} & \textbf{SI $\uparrow$} & \textbf{SD $\downarrow$} & \textbf{CI $\uparrow$} & \textbf{CD $\downarrow$} & \textbf{SAD $\downarrow$} & \textbf{CAD $\downarrow$}  \\ \hline

Conditional Occlusion  &   &   & \textbf{0.781}   & 0.748  & \textbf{0.611}  & \textbf{0.217}    & \textbf{0.155}       & \textbf{0.295}    & \textbf{0.801}   & 0.716   & \textbf{0.647}  & \textbf{0.212}   & \textbf{0.161}        & \textbf{0.285}        \\ 
Pairwise Occlusion   &      &        & 0.757     & \textbf{0.632}             & 0.602              & 0.278               & 0.371     & 0.612    & 0.773   & \textbf{0.640}   & 0.643  & 0.274  & 0.347   & 0.564        \\  \hline
Deep Similarity \cite{Stylianou2019VisualizingDS}  &     &            & 0.794             & 0.770             & \textbf{0.631}               & 0.229               & 0.137                     & 0.297         & \textbf{0.806}   & 0.742   & \textbf{0.663}  & 0.213   & 0.166  & 0.302              \\
Baseline-Grad-CAM      &    &                   & 0.781             & 0.764             & 0.623               & 0.236              & 0.135        & \textbf{0.287}      & 0.798  & 0.737   & 0.661  & 0.211    & 0.166       & \textbf{0.298}                       \\ 
Int-CAM      & Mean  &   $\xmark$     & 0.778             & \textbf{0.739}             & 0.613               & 0.237              & 0.157    & 0.341  & 0.794   & \textbf{0.721}  & 0.651  & 0.212 & 0.200  & 0.367    \\ 
Int-CAM    & Mean   & $\checkmark$            & 0.789             & 0.765             & 0.616               & \textbf{0.228}               & 0.151     & 0.332   & 0.798 & 0.743  & 0.650   & \textbf{0.210}  & 0.187   & 0.343       \\ 
Int-CAM  & Max  &    $\xmark$             & 0.786             & 0.761             & 0.625               & 0.233               & 0.157      & 0.411   & 0.804 & 0.730  & 0.661   & 0.211  & 0.188       & 0.435       \\
Int-CAM   & Max  & $\checkmark$               & \textbf{0.795}             & 0.778             & 0.627              & \textbf{0.228}     & 0.149    & 0.414    & 0.805 & 0.749  & 0.658   & 0.212  & 0.174       & 0.421   \\
Int-CAM    & Attn  &   $\xmark$    & 0.785             & 0.773             & 0.617               & 0.239     & 0.149   & 0.567 & 0.802 & 0.739  & 0.652   & 0.220  & 0.166 & 0.602 \\ 
Int-CAM  & Attn  & $\checkmark$         & 0.793      & 0.782    & 0.620               & 0.236      & \textbf{0.127}    & 0.579   & 0.803 & 0.751  & 0.651   & 0.223 & \textbf{0.151}   & 0.598     \\ \hline
Input x Gradient      &  &               &0.895 &0.895 &0.558 &0.102 &0.048 &0.931  &0.888 &0.839 &0.582 &0.102 &0.091 &0.937 \\
Input x Gradient (G)    &  &       &0.899 &0.906 &0.678 &0.075 &0.062 &0.813  &0.904 &0.867 &0.726 &\textbf{0.069} &0.082 &0.798  \\
Smooth-Grad  &  &    &0.894 &0.913 &0.606 &\textbf{0.072} &0.073 &0.853  &0.884 &0.854 &0.639 &\textbf{0.069} &0.119 &0.832  \\
Smooth-Grad (G) &  &  &0.882 &0.897 &0.658 &0.101 &0.078 &\textbf{0.748}  &0.883 &0.847 &0.693 &0.112 &0.106 &\textbf{0.735}  \\
Avg. Transforms   &  &       &0.898 &0.904 &0.557 &0.100 &\textbf{0.043} &0.929  &0.892 &0.849 &0.581 &0.100 &0.080 &0.935  \\
Avg. Transforms (G)  &  &   &\textbf{0.901} &0.915 &\textbf{0.680} &0.075 &0.050 &0.807  &\textbf{0.907} &0.877 &\textbf{0.728} &\textbf{0.069} &\textbf{0.069} &0.795  \\
Avg. Transforms (N)  &  &   &0.855 &\textbf{0.843} &0.569 &0.089 &0.115 &0.911 &0.843 &\textbf{0.779} &0.604 &0.092 &0.154 &0.912  \\ \hline
\end{tabular}
}
\label{tab1}
\end{table*}

\begin{table}[]
\centering
\caption{Evaluation scores on perturbation, activation and saliency-based methods for Barlow Twins}
\scalebox{0.7}{
\begin{tabular}{c|c|c|c|c|c|c|c|c}
\hline
& AR & GI & \textbf{SI $\uparrow$} & \textbf{SD $\downarrow$} & \textbf{CI $\uparrow$} & \textbf{CD $\downarrow$} & \textbf{SAD $\downarrow$} & \textbf{CAD $\downarrow$} \\ \hline

Conditional Occlusion  &   &    &\textbf{0.655} &\textbf{0.421} &\textbf{0.540} &\textbf{0.180} &\textbf{0.517} &\textbf{0.549} \\
Pairwise Occlusion            &   &    &0.652 &0.497 &0.537 &0.188 &0.549 &0.682 \\  \hline
Deep Similarity  \cite{Stylianou2019VisualizingDS}              &   &    &\textbf{0.706} &0.484 &\textbf{0.578} &0.177 &\textbf{0.362} &\textbf{0.415} \\   
Baseline-Grad-CAM      &    &   &0.688 &0.489 &0.571 &0.180 &0.386 &0.433 \\
Int-CAM  & Mean  & $\xmark$      &0.680 &\textbf{0.473} &0.567 &\textbf{0.169} &0.416 &0.447 \\  
Int-CAM  & Mean  & $\checkmark$  &0.693 &0.490 &0.571 &\textbf{0.169} &0.381 &0.427 \\ 
Int-CAM  & Max  &   $\xmark$     &0.693 &0.479 &0.574 &0.170 &0.437 &0.531 \\ 
Int-CAM  & Max  & $\checkmark$   &0.701 &0.490 &0.577 &0.171 &0.398 &0.502 \\ 
Int-CAM  & Attn  &  $\xmark$     &0.693 &0.494 &0.567 &0.172 &0.458 &0.694 \\ 
Int-CAM  & Attn  & $\checkmark$  &0.699 &0.498 &0.571 &0.174 &0.451 &0.679 \\  \hline
Input x Gradient   &   &  &0.718 &0.577 &0.495 &0.119 &0.472 &0.981    \\ 
Input x Gradient (G) &   & &0.825 &0.599 &0.703 &\textbf{0.069} &0.471 &0.938     \\   
Smooth-Grad     &   &      &0.749 &0.572 &0.540 &0.089 &0.460 &0.972    \\  
Smooth-Grad (G)   &   &    &\textbf{0.832} &0.606 &\textbf{0.713} &0.077 &0.456 &\textbf{0.928}     \\ 
Avg. Transforms  &   &     &0.724 &0.585 &0.494 &0.117 &\textbf{0.437} &0.982    \\ 
Avg. Transforms (G) &   &  &0.829 &0.617 &0.705 &0.070 &0.453 &0.937    \\
Avg. Transforms (N) &   &  &0.714 &\textbf{0.548} &0.504 &0.111 &0.505 &0.976   \\ \hline
\end{tabular}
}
\label{tab3}
\end{table}

\begin{table}[]
\centering
\caption{Evaluation scores on perturbation, activation and saliency-based methods for SimSiam}
\scalebox{0.7}{
\begin{tabular}{c|c|c|c|c|c|c|c|c}
\hline
& AR & GI & \textbf{SI $\uparrow$} & \textbf{SD $\downarrow$} & \textbf{CI $\uparrow$} & \textbf{CD $\downarrow$} & \textbf{SAD $\downarrow$} & \textbf{CAD $\downarrow$} \\ \hline

Conditional Occlusion  &     &     &\textbf{0.809} &0.549 &\textbf{0.640} &\textbf{0.279} &\textbf{0.286} &\textbf{0.380} \\  
Pairwise Occlusion     &     &     &0.723 &\textbf{0.459} &0.600 &0.357 &0.798 &0.593 \\   \hline
Deep Similarity  \cite{Stylianou2019VisualizingDS}   &     &     &0.827 &0.606 &\textbf{0.663} &0.274 &\textbf{0.221} &\textbf{0.312} \\ 
Baseline-Grad-CAM     &     &     &0.816 &0.585 &0.659 &0.279 &0.243 &0.337 \\  
Int-CAM  & Mean  & $\xmark$      &0.815 &\textbf{0.565} &0.654 &0.268 &0.280 &0.382 \\    
Int-CAM  & Mean  & $\checkmark$  &\textbf{0.828} &0.606 &0.655 &\textbf{0.263} &0.236 &0.351 \\   
Int-CAM  & Max  &   $\xmark$     &0.822 &0.575 &0.662 &0.269 &0.294 &0.433 \\  
Int-CAM  & Max  & $\checkmark$   &0.827 &0.613 &0.662 &0.266 &0.252 &0.411 \\ 
Int-CAM  & Attn  &  $\xmark$     &0.816 &0.566 &0.654 &0.267 &0.280 &0.385 \\  
Int-CAM  & Attn  & $\checkmark$  &\textbf{0.828} &0.606 &0.656 &\textbf{0.263} &0.237 &0.355 \\   \hline
Input x Gradient   &     &      &0.885 &0.780 &0.595 &0.192 &0.262 &0.966 \\  
Input x Gradient (G) &     &      &0.925 &0.820 &0.809 &\textbf{0.124} &0.242 &0.876 \\  
Smooth-Grad   &     &      &0.897 &0.787 &0.639 &0.154 &0.247 &0.949 \\  
Smooth-Grad (G)   &     &      &0.924 &0.819 &\textbf{0.812} &0.132 &0.232 &\textbf{0.867} \\ 
Avg. Transforms  &     &      &0.888 &0.786 &0.595 &0.189 &0.234 &0.965 \\  
Avg. Transforms (G) &     &      &\textbf{0.926} &0.830 &0.810 &0.126 &\textbf{0.227} &0.874 \\  
Avg. Transforms (N) &     &      &0.875 &\textbf{0.747} &0.602 &0.178 &0.300 &0.963 \\  \hline
\end{tabular}
}
\label{tab_evalsimsiam}
\end{table}

\begin{table}[]
\centering
\caption{Evaluation scores on perturbation, activation and saliency-based methods for MoCov3 ViT-Base}
\scalebox{0.7}{
\begin{tabular}{c|c|c|c|c|c|c|c|c}
\hline
& AR & GI & \textbf{SI $\uparrow$} & \textbf{SD $\downarrow$} & \textbf{CI $\uparrow$} & \textbf{CD $\downarrow$} & \textbf{SAD $\downarrow$} & \textbf{CAD $\downarrow$} \\ \hline

Conditional Occlusion  &     &     &\textbf{0.809} &\textbf{0.622} &\textbf{0.714} &\textbf{0.280} &0.483 &0.533  \\
Pairwise Occlusion    &     &     &0.807 &0.685 &0.706 &0.435 &\textbf{0.378} &\textbf{0.439}  \\    \hline
Deep Similarity  \cite{Stylianou2019VisualizingDS}   &     &     &0.782 &0.718 &0.627 &0.414 &\textbf{0.260} &\textbf{0.270}  \\
Baseline-Grad-CAM     &     &     &\textbf{0.805} &0.667 &\textbf{0.690} &\textbf{0.376} &0.393 &0.603  \\
Int-CAM  & Mean  & $\xmark$      &0.795 &\textbf{0.653} &0.670 &0.389 &0.476 &0.800  \\  
Int-CAM  & Mean  & $\checkmark$  &0.787 &0.716 &0.599 &0.478 &0.314 &0.410  \\  
Int-CAM  & Max  &   $\xmark$      &0.799 &0.666 &0.683 &0.382 &0.422 &0.670  \\  
Int-CAM  & Max  & $\checkmark$   &0.787 &0.714 &0.606 &0.467 &0.296 &0.350  \\ 
Int-CAM  & Attn  &  $\xmark$     &0.794 &0.656 &0.670 &0.388 &0.467 &0.780  \\   
Int-CAM  & Attn  & $\checkmark$  &0.787 &0.715 &0.606 &0.468 &0.314 &0.422  \\    \hline
Input x Gradient  &     &     &0.902 &0.760 &0.746 &0.363 &0.280 &0.904  \\  
Input x Gradient (G) &     &     &0.901 &0.749 &0.752 &0.341 &0.331 &0.889  \\  
Smooth-Grad   &     &     &0.909 &0.750 &0.762 &0.309 &0.284 &0.833  \\ 
Smooth-Grad (G)   &     &     &\textbf{0.911} &0.749 &\textbf{0.769} &\textbf{0.295} &0.301 &\textbf{0.817}  \\ 
Avg. Transforms  &     &     &0.904 &0.763 &0.747 &0.359 &\textbf{0.259} &0.903  \\  
Avg. Transforms (G) &     &     &0.902 &0.755 &0.752 &0.339 &0.307 &0.887  \\ 
Avg. Transforms (N) &     &     &0.904 &\textbf{0.733} &0.748 &0.348 &0.345 &0.886   \\  \hline
\end{tabular}
}
\label{tab_evalmocov3}
\end{table}

\begin{table}[]
\centering
\caption{Maximum Sensitivity evaluation on saliency-based methods. Lower is better.}
\scalebox{0.9}{
\begin{tabular}{c|c|c|c|c}
\hline
& SimCLR(1X) & SimCLR(2X) & SimSiam & MoCov3 \\ \hline
Input x Gradient     &1.386    &1.799     &2.306   &3.121   \\
Input x Gradient (G) &1.402    &2.178     &3.168   &3.231   \\ 
Smooth-Grad          &1.208    &1.257     &2.534   &3.317   \\  
Smooth-Grad (G)      &1.652    &1.780     &1.950   &2.700  \\ 
Avg. Transforms      &1.581    &1.910     &1.876   &2.772   \\ 
Avg. Transforms (G)  &1.448    &2.301     &\textbf{1.711}   &2.270  \\ 
Avg. Transforms (N)  &\textbf{0.923}    &\textbf{0.976}     &1.907   &\textbf{2.186}   \\  \hline
\end{tabular}
}
\label{tab_maxsensitivity}
\end{table}

\section{Evaluation Metrics}
\label{sec:eval}
In this section, we first re-visit metrics commonly used to evaluate single explanations of image classification systems, and then show how we build upon these metrics and adapt them to suit the evaluation of pairs of explanations. 

\subsection{Metrics for Evaluating Explanations of Image Classification Systems}
\label{sec:ExistingMetrics}
We consider the following metrics: insertion and deletion \cite{Petsiuk2018RISERI}, average drop \cite{Chattopadhyay2018GradCAMGG}, maximum sensitivity \cite{Yeh2019HowSA} and sanity checks \cite{Adebayo2018SanityCF}. In the insertion and deletion metric, we start from a baseline (a highly blurred image for insertion and the full image for deletion) and gradually add (or remove) image pixels starting from the pixels deemed most important by the explanation algorithm. The output score curve is then plotted as a function of the pixels being added or removed and the Area Under the Curve (AUC) is computed. A sharp increase (or decrease) in this curve generally reflects that the explanation is faithful and complete. The Average Drop~\cite{Chattopadhyay2018GradCAMGG} was developed to measure how the model’s confidence drops when providing only the explanation map regions as input, with lower scores being better. The Maximum sensitivity~\cite{Yeh2019HowSA} measures to which extent the explanation changes when the input is slightly perturbed with noise. Sanity Check~\cite{Adebayo2018SanityCF} randomizes each layer of the model, starting from the last layer up to the first layer. At each step of randomization, the explanation map is visualized to ensure that the explainability methods do not act as edge detectors, and that they rely on the model parameters for a prediction. 
Lastly, we refrain from evaluating the explanations of contrastive models with localization since this is only valid for evaluating explanations of models trained for image classification. As a result, the model must be able to localize the object adequately in order to classify the object present in the image. However, this is not the case for contrastive learning paradigms where the objective is to learn similarity. In fact, it is very rare to see the explanation of a similarity model covering the full object from both images. Instead, we expect to see associations between the two images that convey meaningful and relevant concepts about the object. 

\subsection{Adaptation to Pairs of Explanations}
\label{sec:AdaptedMetrics}
Unlike image classification systems which yield a single explanation, the output of our explanation algorithms produce pairs of explanations. We therefore expand the existing metrics discussed in Section~\ref{sec:ExistingMetrics} to suit the evaluation of pairs of explanations. We perform this under two settings: a simultaneous and a conditional setting, each of which is described below. 

\subsubsection{Simultaneous Evaluation} In this setting, we simultaneously evaluate both explanation maps. In the case of insertion and deletion, we add or remove $L$ pixels from both the first and second image \textit{together} at each step of the iteration process, starting from the most important pixels in the explanation map. We then measure the increase or drop in the similarity score between the two altered images with $L$ added or removed pixels each (in our experiments we set $L$ to 224). Applying this setting on the insertion and deletion metrics yields simultaneous insertion (SI) and simultaneous deletion (SD), respectively. Similarly, applying it on the average drop metric yields simultaneous average drop (SAD). However, this approach assumes a \textit{correspondence} between the two explanation maps, meaning that a point in the first image must correspond to a point in the second image. In that case, the two explanation maps are dependant on each other and treated as a whole. This is made evident by observing a gradual, less sharp increase or decrease in the output score curve as pixels are added or removed. This mainly indicates that the similarity can still be computed from the context around the two images. This motivates us to propose another variant of evaluation, namely \textit{Conditional Evaluation}. 

\subsubsection{Conditional Evaluation} In this setting, we evaluate one explanation map conditioned on the other image being fully observed. In the case of insertion and deletion, we add or remove $L$ pixels from the first image while holding the other image constant. At each step, we measure the increase or drop in the similarity score between the altered image with $L$ added or removed pixels, and the original unaltered second image. The operation is done for each of the two images and the results are averaged. Applying this setting on the insertion and deletion metrics yields conditional insertion (CI) and conditional deletion (CD), respectively. Similarly, applying it on the average drop metric yields conditional average drop (CAD). The main drawback of this setting is that it doubles the evaluation time for each image in the evaluation set. 

In what follows, we assess all our techniques by subjecting them to simultaneous and conditional evaluations across all six metrics. 

\begin{figure}
    \centering
    \includegraphics[width=0.5\textwidth]{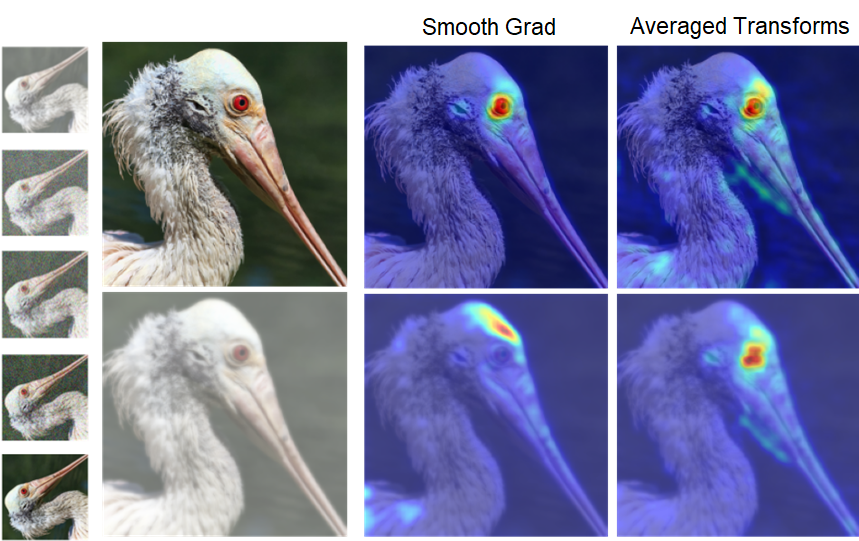}
    \caption{Qualitative examples of Averaged Transforms compared with Smooth-Grad. On the left, we show the augmentation dissection process of gradually varying the transformation strength of the first image until it reaches the second image.}
    \label{avgtransforms}
\end{figure}

\begin{figure*}
    \centering
    \includegraphics[width=0.95\textwidth]{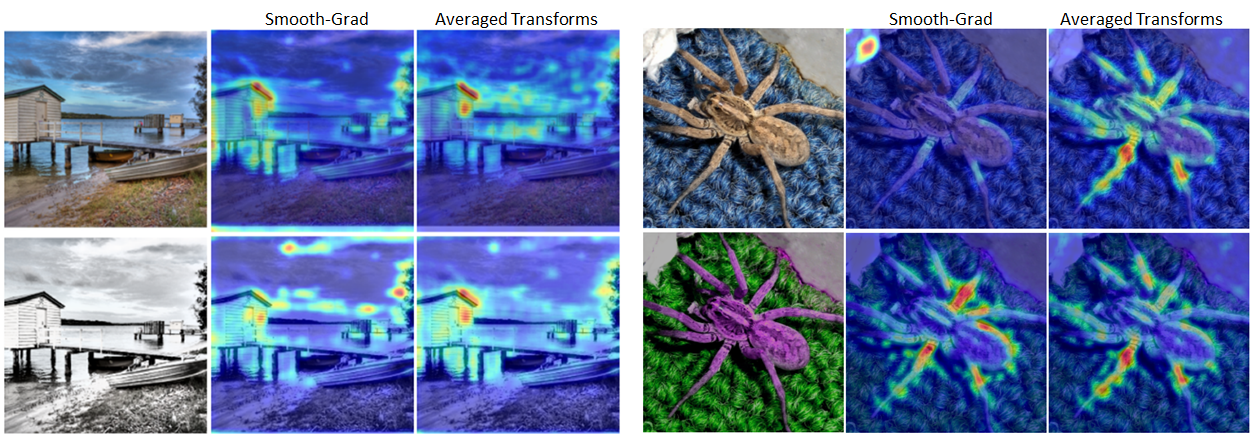}
    \caption{Qualitative examples of Averaged Transforms compared with Smooth-Grad, for grayscale and color jitter transformations.}
    \label{avgtransforms2}
\end{figure*}

\begin{figure}
    \centering
    \includegraphics[width=0.5\textwidth]{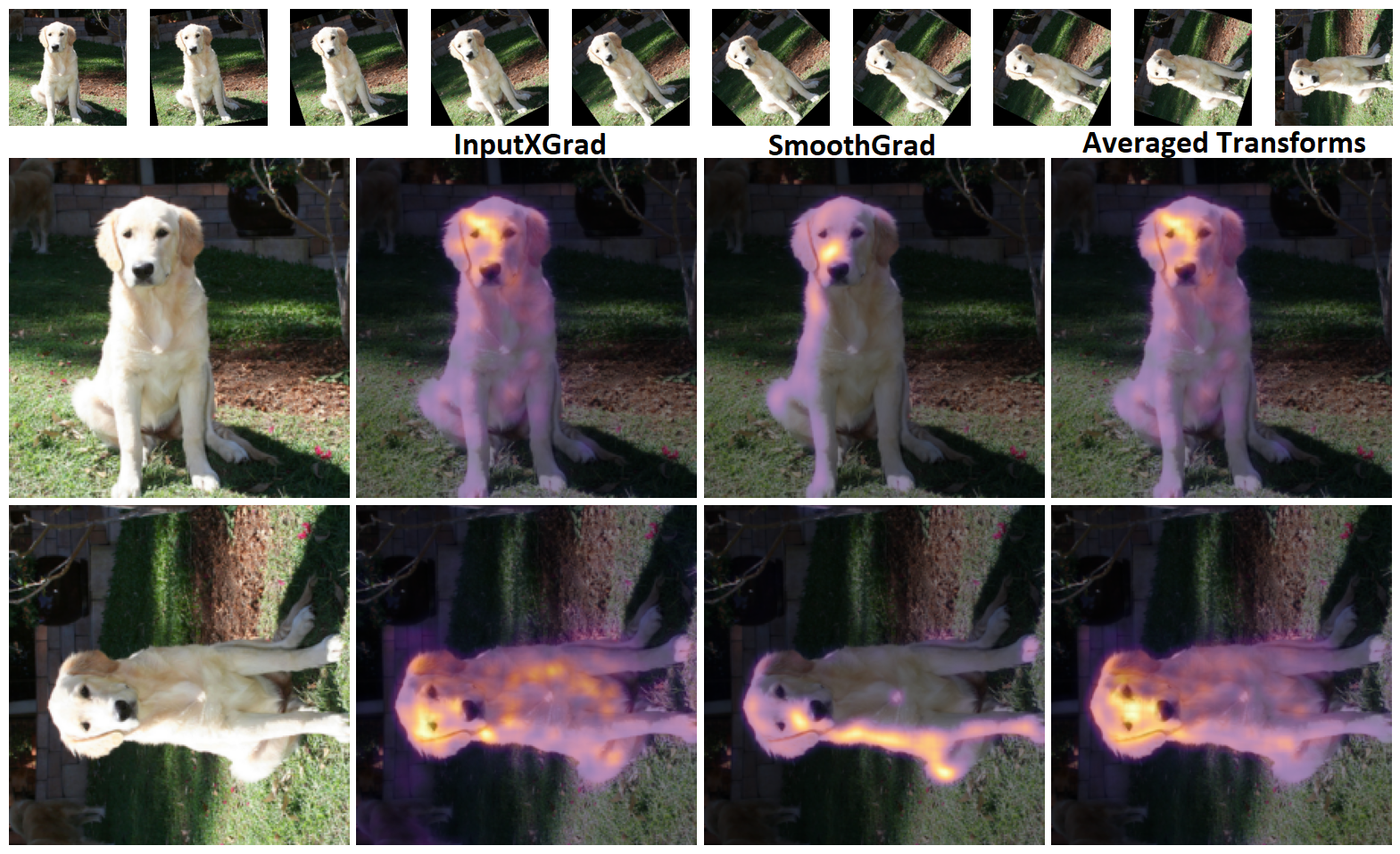}
    \caption{Qualitative examples of Averaged Transforms compared to InputXGrad and Smooth-Grad, for a 90 degree rotation transformation. On top, we show the augmentation dissection process.}
    \label{avg_transf_rotation}
\end{figure}

\begin{figure*}
    \centering
    \includegraphics[width=0.95\textwidth]{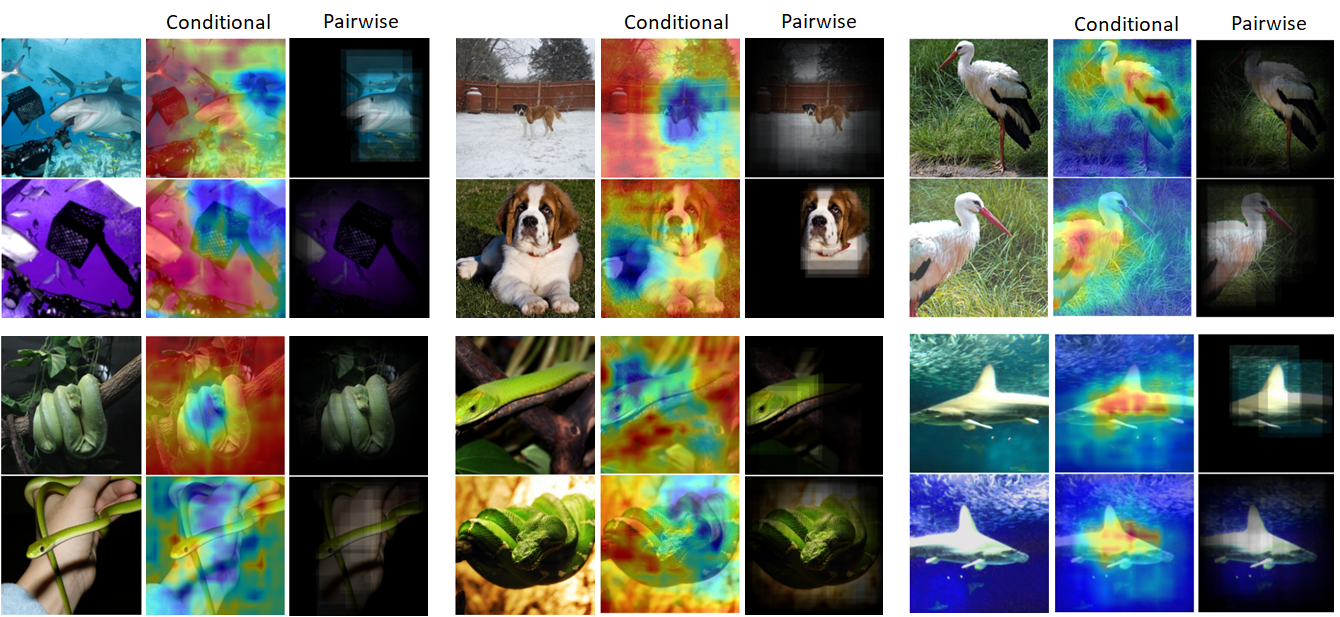}
    \caption{6 qualitative examples of Perturbation-based methods. Our methods are not limited to explaining similarity on augmented pairs. We can also apply them to explain similarity of different images from the same class. The bottom example in the first column and the top and bottom examples in the second column present qualitative results on different images of the same class. The other qualitative examples (the top example in the first column and the two examples in the last column) are for augments of an image. Red signifies more important regions while blue signifies less important regions.}
    \label{occlusion}
\end{figure*}

\begin{figure*}
    \centering
    \includegraphics[width=0.9\textwidth]{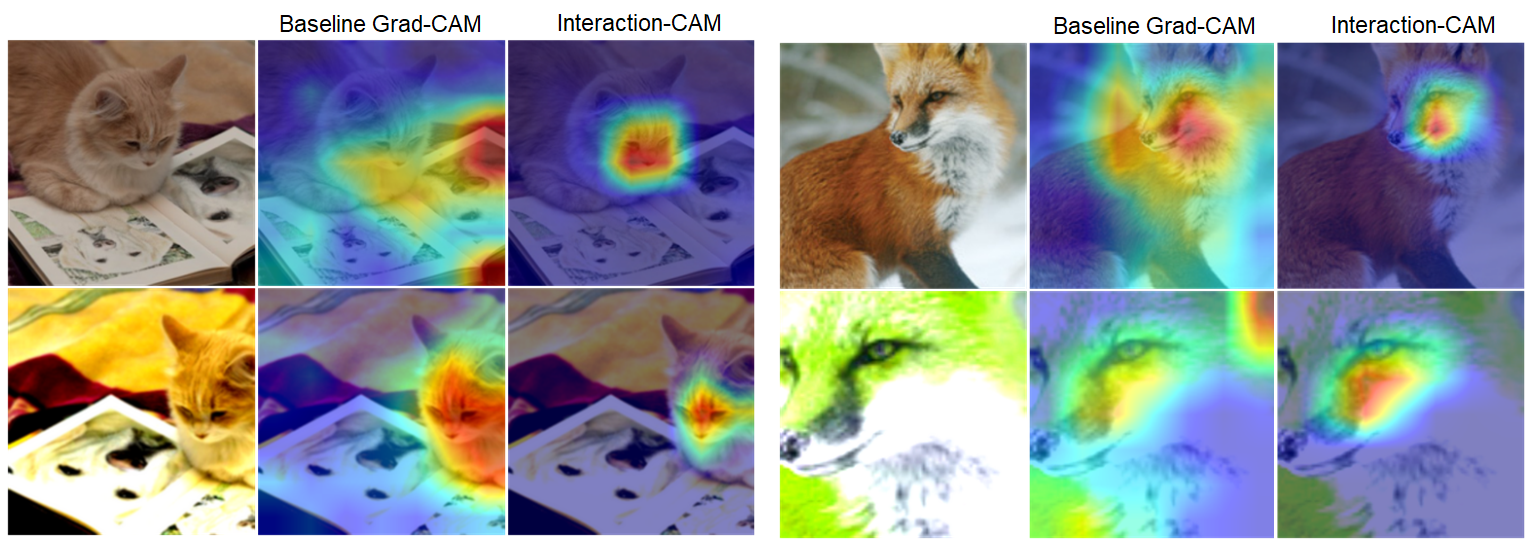}
    \caption{Qualitative examples of the proposed Interaction-CAM compared to Baseline-Grad-CAM.}
    \label{cam}
\end{figure*}

\begin{figure*}[t]
    \centering
    \includegraphics[width=0.84\textwidth]{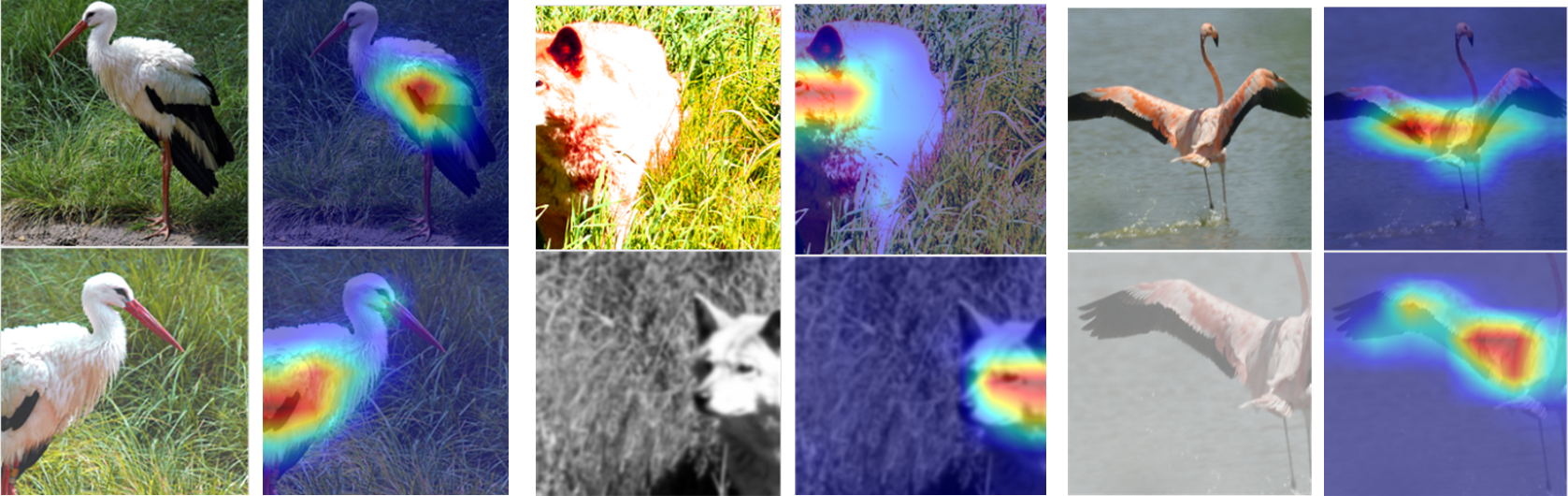}
    \caption{Examples of meaningful concepts learned by the model and detected by Interaction-CAM, where the same concepts related to a particular object are learned. The positions of these meaningful concepts are aligned between the two images (\textit{e.g.,} the wolf’s eye, the flamingo wing}).
    \label{strong_concepts2}
\end{figure*}

\begin{figure}[t]
    \centering
    \includegraphics[width=0.5\textwidth]{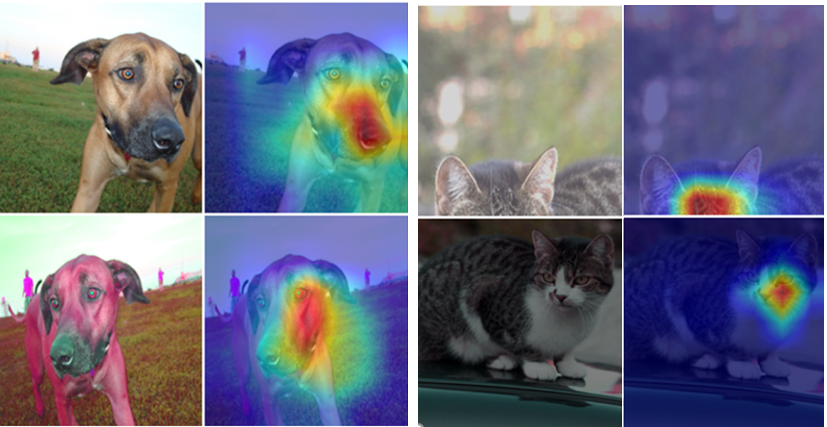}
    \caption{Examples of meaningful concepts learned by the model and detected by Interaction-CAM, where different semantic concepts associated to the same object are learned (\textit{e.g.,} different parts of the dog’s and cat’s faces).}
    \label{strong_concepts1}
\end{figure}

\begin{figure*}[t]
    \centering
    \includegraphics[width=0.9\textwidth]{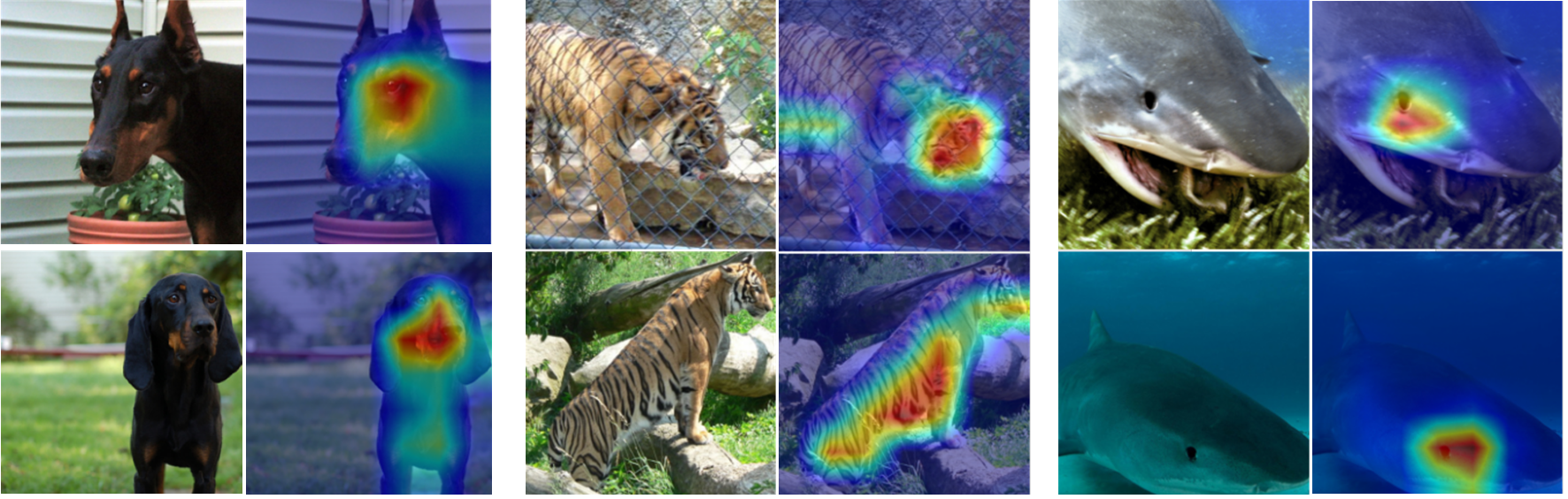}
    \caption{Interaction-CAM applied on two different images from the same class.}
    \label{same_class}
\end{figure*}

\begin{figure*}
    \centering
    \includegraphics[width=0.9\textwidth]{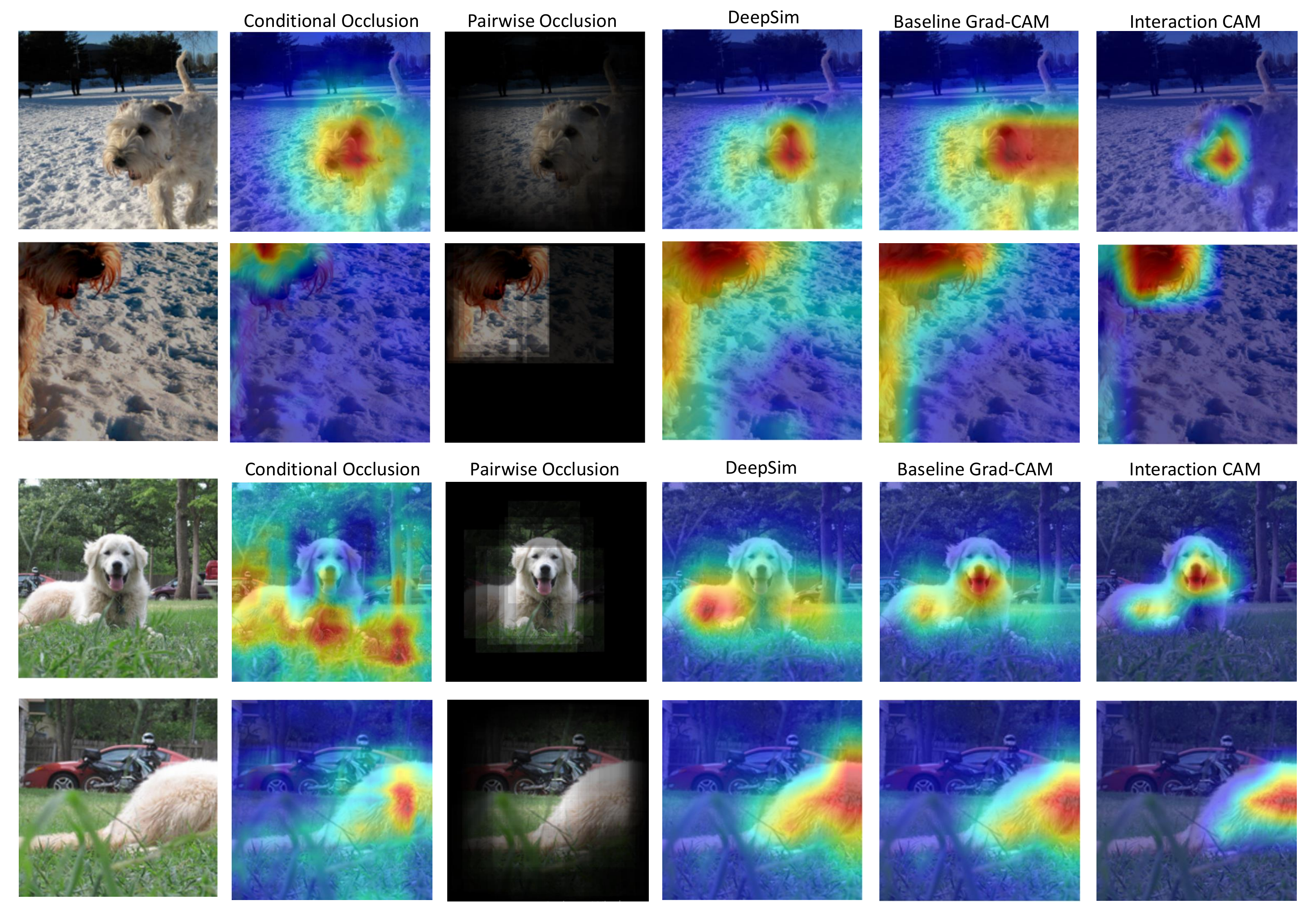}
    \caption{A visualization of perturbation and activation-based methods applied on the same image pairs. DeepSim is the technique from \cite{Stylianou2019VisualizingDS}}
    \label{all_vizs}
\end{figure*}

\section{Experiments}
\label{sec:exp}
In this section, we conduct a qualitative and a quantitative analysis (using the six proposed metrics) of the proposed explanation methods. 

\subsection{Experimental Setup}
 We randomly choose a subset of $2000$ images from the ImageNet ILSVRC-2012 \cite{Russakovsky2015ImageNetLS} validation set. We experiment with five different models: SimCLRv2 1$\times$, SimCLRv2 2$\times$ \cite{chen2020big}, Barlow Twins \cite{Zbontar2021BarlowTS}, SimSiam \cite{Chen2021ExploringSS} and MoCov3 \cite{Chen2021AnES}. All methods are evaluated using the \textit{cosine similarity} measure. SimCLRv2 is a contrastive model which utilizes data augmentation to obtain two views of the same image and learns a similarity task; namely, augments of the same image should similar, while augments from different images within the same batch of samples should be dissimilar. It consists of a vision backbone (\textit{e.g.,} CNN \cite{LeCun1998GradientbasedLA} or Transformer \cite{Vaswani2017AttentionIA}) followed by a projection head that learns invariance. SimCLRv2 2$\times$ doubles the width of the CNN, and has shown to perform much better than SimCLRv2 1$\times$. SimSiam \cite{Chen2021ExploringSS} is a contrastive model that eliminates the need of negative pairs through methodologies such as the stop-gradient mechanism, a predictor network, and momentum-based encoding. MoCov3 is another contrastive model that is similar to SimCLRv2, but uses a Vision Transformer as the backbone and uses separate models to encode each view of the image: the query encoder and the key encoder. The query encoder has an extra prediction head, and the key encoder is updated by a moving average of the query encoder. Barlow Twins is a self-supervised model which is inspired by Principal Component Analysis. Barlow Twins estimates a covariance matrix which is trained to be as close to the identity matrix as possible. As in SimCLR, data augmentation is used to obtain two different views of the same image. It is worth noting that the pretext task of Barlow Twins is not contrastive; however, it still requires two inputs in order to learn. As a result, we can use our methods to analyze the similarity aspect of this model as well.

\subsection{Qualitative Analysis}
We initially present visual examples of our methods. In Figs. \ref{avgtransforms} and \ref{avgtransforms2}, we show how one of the baseline methods (Smooth-Grad) highlights incorrect corresponding regions as a result of considering only one level of augmentation, which can be misleading. Our proposed Averaged Transforms method mitigates this issue by averaging over all augmentation levels, thereby diminishing irrelevant parts. We present further qualitative results for Averaged Transforms using a 90-degree rotation transformation, as shown in Fig.~\ref{avg_transf_rotation}. The top row shows the dissected rotation steps. We compare the results with InputXGrad and SmoothGrad in the second and third column, respectively. It is worth highlighting that while the rotation transformation is not part of the contrastive learning methods studied, its assessment is still relevant. This is because the contrastive model is expected to be invariant across various transformations, even across those that it has not been trained on, such as rotations. In Fig.~\ref{occlusion}, we observe how in many cases, the baseline Conditional Occlusion deems the background as highly important, due to the problem discussed in Section~\ref{sec:perturbation}. Our proposed Pairwise Occlusion method  alleviates this problem. Particularly, regions identified as unimportant by Conditional Occlusion are sometimes deemed significant in Pairwise Occlusion. Next, we show visual examples of each technique. Figure~\ref{cam} shows the output of Interaction-CAM compared to Baseline-Grad-CAM. In spite of the high similarity score between the two images, the baseline conveys irrelevant concepts. Interaction-CAM concentrates all the energy in similar concepts which jointly contribute to the high similarity output. Moreover, we observe in multiple cases that contrastive models can learn to attend to semantically meaningful concepts, despite being trained in a self-supervised class-free setting. These concepts are often visually salient due to their distinctive textures and geometric patterns. Figure~\ref{strong_concepts2} illustrates this point and shows that the important visual cues are associated to the main subject of the scene, and the locations of the meaningful concepts are correctly mapped between the two images, including the region around the wolf's eye or the back of the flamingo. Figure~\ref{strong_concepts1} also shows how different concepts are highlighted when those are associated to the same semantic object (different parts of the dog's and cat's faces). The ability to learn distinctive visual cues is also illustrated in Fig.~\ref{same_class} where two different images from the same class are fed to the similarity model, showing a strong attribution to distinctive patterns. In Fig.~\ref{all_vizs}, we show qualitative results from our perturbation- and activation-based approaches, with all methods applied on the same pair of images. This shows how different approaches interpret the same image. Activation-based methods are also compared with DeepSim \cite{Stylianou2019VisualizingDS}. As evident from the Figure, most methods are in consensus regarding what they identify as important. 

\subsection{Quantitative Analysis}
We will now present the evaluation scores for SimCLRv2~1$\times$, SimCLRv2 2$\times$, Barlow Twins, SimSiam and MoCov3 across all metrics proposed in Section~\ref{sec:AdaptedMetrics}. Subsequently, we will provide a summary of our findings through an additional analysis to enhance the comprehensibility and clarity of the results. Evaluation scores from the SimCLRv2 (1$\times$, 2$\times$) contrastive models \cite{chen2020big} are presented in Table \ref{tab1}. Barlow Twins model results are displayed in Table \ref{tab3}. SimSiam and MoCov3 evaluation scores are shown in Tables~\ref{tab_evalsimsiam}~and~\ref{tab_evalmocov3}, respectively. SimCLRv2 (1$\times$), SimCLRv2 (2$\times$), Barlow Twins and SimSiam use a ResNet-50~\cite{He2016DeepRL}  backbone, whereas MoCov3 uses a Vision Transformer backbone (ViT-Base). Since Interaction-CAM belongs to the activation visualization category of explainability, we also compare Interaction-CAM with the Deep Similarity technique \cite{Stylianou2019VisualizingDS} on all metrics. Upon examining the results presented in the tables, it is apparent that the baseline conditional occlusion method outperforms our proposed pairwise occlusion approach on nearly all metrics. This can be attributed to the fact that the pairwise occlusion method tends to generate sparse regions with equal pixel values due to the assignment of the softmax score to all pixels within the erased area. The regions—specifically, the perturbed rectangles—end up with pixels sharing identical values (the softmax score). Yet, our evaluation mechanism assesses based on individual pixels rather than broader regions. As a result, when a whole region is assigned one uniform value, it causes a plateau in the evaluation curve. Therefore, the evaluation curve does not exhibit sharp drops or increases when pixels are added or removed, as is observed in the case of conditional occlusion. Consequently, this leads to a reduction in the AUC. For saliency-based methods, we observe that Averaged Transforms achieves superior performance compared to baseline methods across most metrics. Particularly, Averaged Transforms variants have obtained the highest scores in 5 out of 7 metrics for SimCLRv2 ($1\times$), and 6 out of 7 metrics for SimCLRv2 ($2\times$). We discuss which variant of Averaged Transforms is generally better in Section~\ref{sec:summary}. For activation-based methods, we observe that the proposed Interaction-CAM variants outperform the baseline methods in 4 out of 6 metrics for SimCLRv2 ($1\times$) and SimCLRv2 ($2\times$). We have additionally demonstrated the impact of the gradient interaction term (GI) on all method variants, revealing a notable enhancement in most performance metrics upon its inclusion. Conversely, Interaction-CAM only outperforms the baseline method on the SD and CD metrics for the Barlow Twins model. This difference in performance could be attributed to the fact that Barlow Twins is not trained with contrastive learning. Thus, assessing the model's explanations using the similarity measure may not yield the most optimal results.  In Section~\ref{sec:summary}, we also investigate which variant of Interaction-CAM is generally better. In Table \ref{tab_maxsensitivity}, we report evaluation scores of Averaged Transforms for the maximum sensitivity metric. As shown, our proposed Averaged Transform achieves the lowest sensitivity when compared to all other baseline measures across all contrastive models.

\begin{figure}
    \centering
    \includegraphics[width=0.5\textwidth]{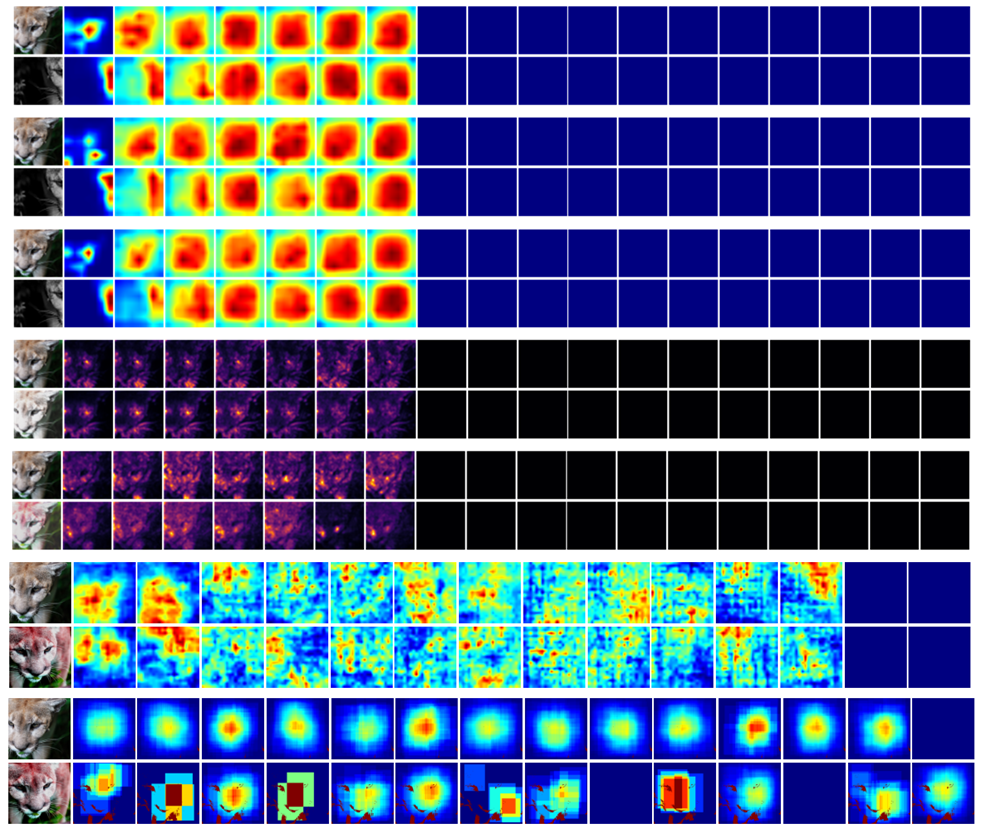}
    \caption{Sanity check for Interaction-CAM (mean), Interaction-CAM (max), Interaction-CAM (attention), Average Transforms (guided backpropagation), Average Transforms (without guided backpropagation), Conditional Occlusion and Pairwise Occlusion, ordered from top to bottom.}
    \label{sanity}
\end{figure}

In Fig.~\ref{sanity}, we perform a sanity check on Interaction-CAM (mean), Interaction-CAM (max), Interaction-CAM (attention), Average Transforms (w/ guided backpropagation), Average Transforms (w/o guided backpropagation), Conditional Occlusion and Pairwise Occlusion, ordered from top to bottom. We start with the original explainability map of the two images (second column) and randomize the weights of the model starting from the top layer, progressively, all the way to the bottom layer. At each randomization step, we run the explainability method and obtain the explanation map. To avoid visual clutter, we only show the results after every three layers. We can see that the explanation completely becomes meaningless as randomization is applied. After a certain layer (e.g., layer 18 in Fig.~\ref{sanity}), the explanation map gets completely zeroed out. Thus, our proposed techniques are sensitive to model parameters and have passed the sanity check.

\subsection{Summary of the Results and Applications}
\label{sec:summary}

\begin{figure*}
    \centering
    \includegraphics[width=\textwidth]{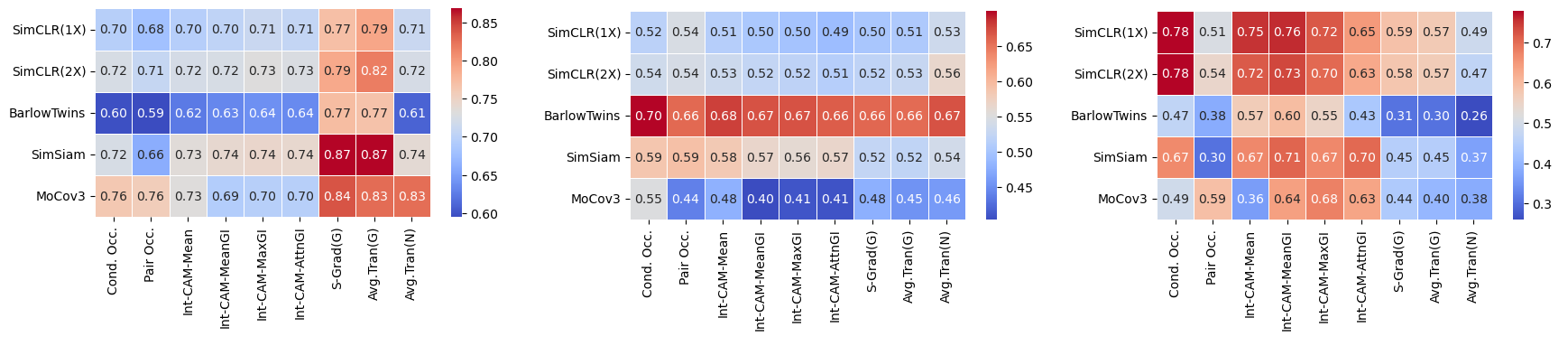}
    \caption{Summary of scores of five self-supervised models for the three groups of metrics (a) SI, CI, (b) SD, CD, and (c) SAD, CAD}
    \label{scores_summary}
\end{figure*}

\begin{figure*}
    \centering
    \includegraphics[width=\textwidth]{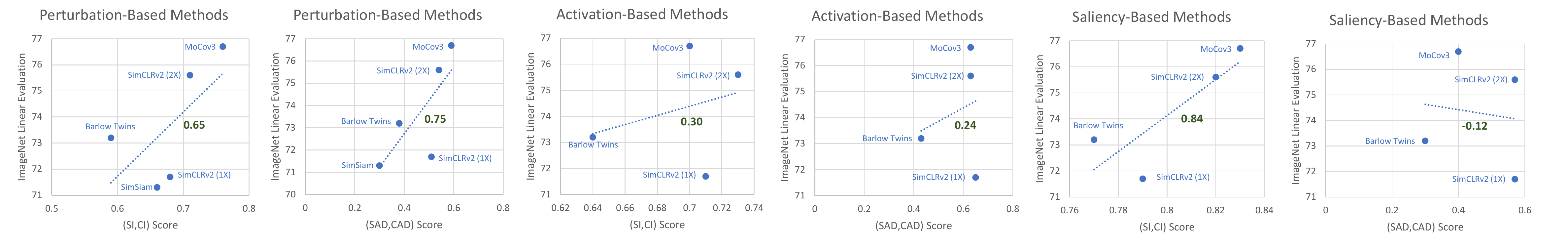}
    \caption{Correlation with downstream tasks on the (SI, CI) and (SAD, CAD) group of metrics shown along with the linear correlation coefficient in green, for perturbation, activation and saliency-based methods. In all plots except the last, a positive correlation is observed between the evaluation scores and downstream task performance.}
    \label{correlation}
\end{figure*}

In Section \ref{sec:exp}, we have presented evaluation scores of our explainability techniques for the five models. In this section,  our aim is to provide a concise and informative summary of the results, thereby facilitating better interpretation. We perform the following analysis to summarize our results. We cluster the evaluation metrics into 3 groups (SI, CI), (SD, CD) and (SAD, CAD), and compute the average score for the first group, and $(1-\text{average})$ for the last two groups, such that a higher score indicates better performance. We thus obtain a single value for each group. It is worth noting that various evaluation metrics are designed to measure different aspects of explanation faithfulness. Deciding which set of metrics to prioritize would depend on what one seeks to assess from the explanation. The first set of metrics, namely (SI, CI), primarily assess the faithfulness of the explanation. Conversely, the last two sets of metrics, (SD, CD) and (SAD, CAD), are primarily focused on capturing the confidence level associated with the explanation. As a result, we refrain from calculating a global average of all six metric. Figure \ref{scores_summary} depicts a heatmap of scores for each of the three groups of metrics for five different models. From Fig. \ref{scores_summary}, we draw the following observations by comparing all methods (row-wise) for each model: For the first group of metrics (SI, CI), Averaged Transforms (G) scores the best on average. For the second and third group of metrics represented by (SD, CD) and (SAD, CAD), respectively, Conditional Occlusion scores best on average. 

Numerous factors play a role in determining the appropriate question that each method answers, i.e., which method to utilize in specific situations. However, we emphasize two key factors that one should consider before making a decision: 1) The user's intent from the explanation, and 2) the computational speed requirements for the explanation. To serve the user with accurate information, it is imperative to determine precisely what the user seeks to comprehend from the contrastive model. For example, Averaged Transforms is a method that can be used to understand how data augmentations and invariance affect the similarity output. On the other hand, perturbation-based method and activation-visualization methods should be used to study the primary causes of the predicted similarity. Another important factor to consider in practical applications is the speed requirements. Table \ref{exec_time} presents the execution time of both the baseline and proposed methods, which is averaged over five separate runs. All experiments were performed on a single NVIDIA RTX 3090 GPU. Our findings show that activation visualization-based methods have the fastest execution time. Therefore, such methods should be favoured if speed is a critical consideration. In general, all methods allow to understand contrastive models, but each differs in its objective and purpose it defines. 

\begin{table}[t]
\centering
\caption{Execution time (in seconds) of the baseline and proposed methods}
\begin{tabular}{|c|c|}
\hline
\textbf{Method}       & \textbf{Time (s)} \\ \hline
Conditional Occlusion & 0.980             \\ \hline
Pairwise Occlusion    & 0.257             \\ \hline
Saliency              & 0.296             \\ \hline
Smooth-Grad           & 2.258             \\ \hline
Avg. Transforms       & 0.411             \\ \hline
Baseline-Grad-CAM     & 0.215             \\ \hline
Int-CAM (w/o GI)      & 0.079             \\ \hline
Int-CAM (w/ GI)       & 0.050             \\ \hline
\end{tabular}
\label{exec_time}
\end{table}

\subsection{Relationship to Downstream Tasks}

\label{sec:downstream}
Although the general focus of this work is to explain contrastive models, we also endeavor to explore the correlation and determine the relationship between the explainability aspect of contrastive self-supervised models and their performance on downstream tasks after finetuning. We use the common image classification task on ImageNet as the downstream task in our experiments. Figure~\ref{correlation} displays the correlation plots of perturbation, activation, and saliency-based methods, along with their corresponding correlation coefficients. The x-axis denotes the explanation score acquired using the same procedure outlined in Section~\ref{sec:summary}, for both the (SI, CI) and (SAD, CAD)  groups of metrics. The $y$-axis denotes the ImageNet linear probe evaluation accuracy. In almost all plots, a positive correlation is observed between the evaluation scores and downstream task performance. Note that for all methods other than perturbation-based ones, we have excluded SimSiam because it was identified as an outlier, which interfered with a correct calculation of the correlation coefficient. Among all family of methods, it can be observed that perturbation-based are mostly correlated with downstream task performance for both metric groups, with a correlation coefficient of 0.75 and 0.65 for (SI, CI) and (SAD, CAD), respectively. We recognize that analyzing only five models may not provide an accurate result, and examining a larger number of models would yield a more precise outcome.

\section{Exploration and Insights}
\label{sec:findings}

\begin{figure}[t]
    \centering
    \includegraphics[width=0.4\textwidth]{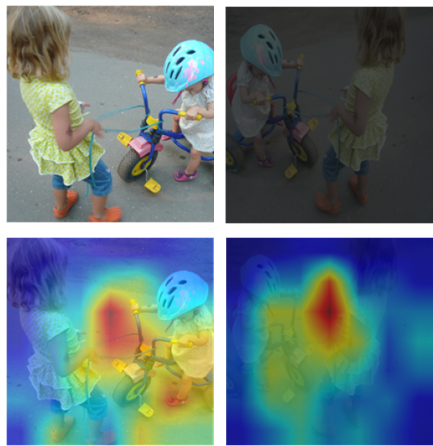}
    \caption{By using Interaction-CAM, we find examples of models which focus on meaningless concepts, despite the same image content being present in both images.}
    \label{meaningless1}
\end{figure}

\begin{figure}[t]
    \centering
    \includegraphics[width=0.5\textwidth]{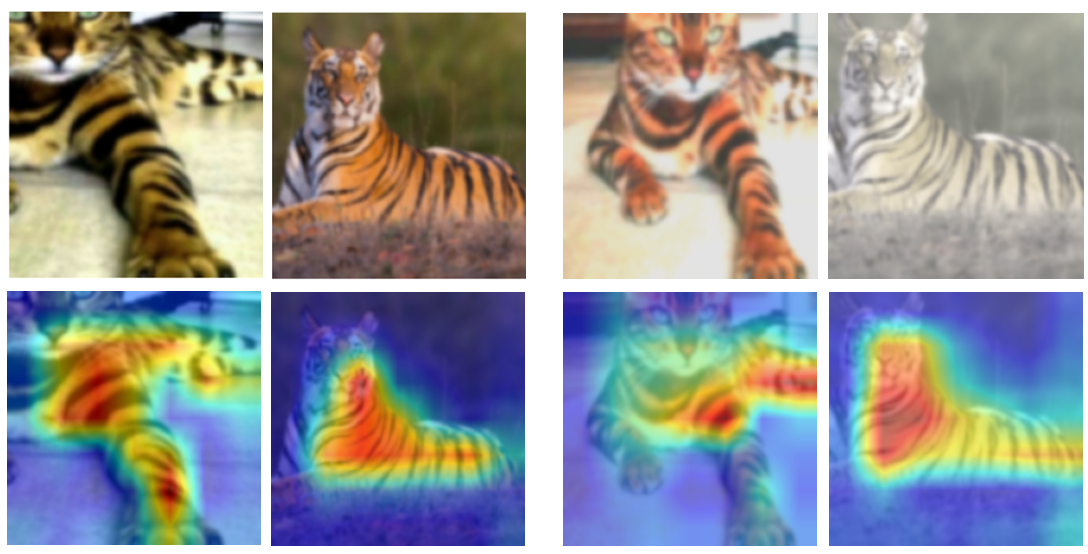}
    \caption{By utilizing Interaction-CAM, we find that a model predicts an image of a "tiger cat" and a "tiger" are similar because of the "stripes" texture, despite the fact that the two images are different. The similarity scores for each of the image pairs (from left to right) are 0.70 and 0.71. }
    \label{texture}
\end{figure}

\begin{figure*}[t]
    \centering
    \includegraphics[width=0.9\textwidth]{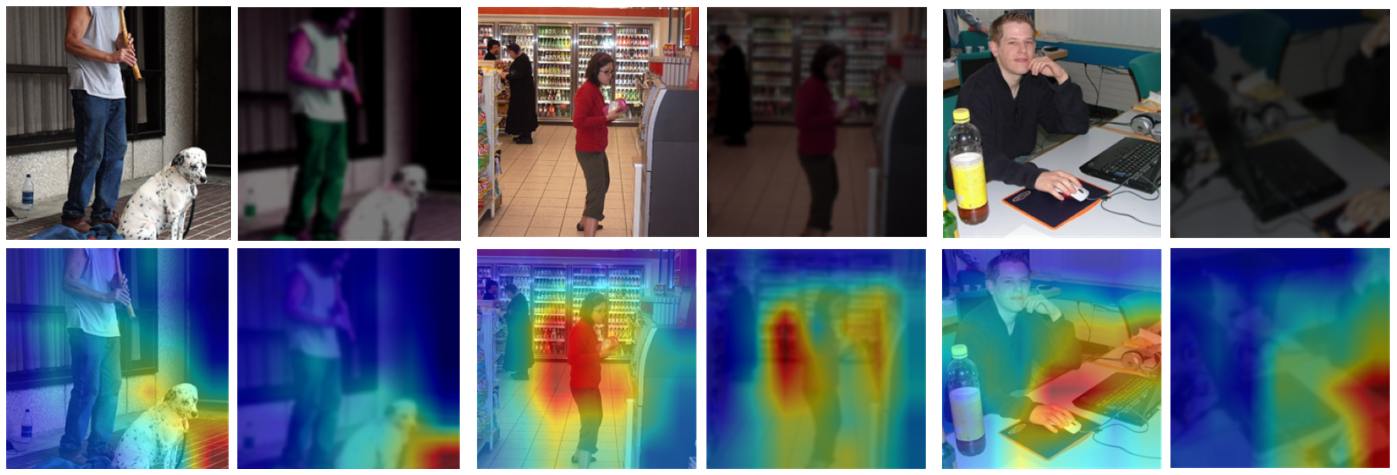}
    \caption{Pairs of images with a low similarity score (<0.5), despite the two images containing the same content. By utilizing Interaction-CAM, we identify that the combination of the \textit{blur and blackening} harms the performance as the model is not able to learn any meaningful concepts.}
    \label{which_augment}
\end{figure*}

\begin{table}[]
\centering
\caption{Explanation scores for Averaged Transforms using several data augmentation techniques.}
\scalebox{0.9}{
\begin{tabular}{|c|cccccc|}
\hline
\textbf{Transformation} & \multicolumn{1}{c|}{\textbf{SI $\uparrow$}}    & \multicolumn{1}{c|}{\textbf{SD $\downarrow$}}    & \multicolumn{1}{c|}{\textbf{SAD $\downarrow$}}   & \multicolumn{1}{c|}{\textbf{SI $\uparrow$}}    & \multicolumn{1}{c|}{\textbf{SD $\downarrow$}}    & \textbf{SAD $\downarrow$}   \\ \hline
                        & \multicolumn{3}{c|}{Non-Guided}                                                                                 & \multicolumn{3}{c|}{Guided}                                                                \\ \hline
                        & \multicolumn{6}{c|}{\textit{SimCLRv2 (1X)}}                                                                                                                                                                  \\ \hline
Color Jitter           & \multicolumn{1}{c|}{0.945}          & \multicolumn{1}{c|}{0.907}          & \multicolumn{1}{c|}{0.052}          & \multicolumn{1}{c|}{0.961}          & \multicolumn{1}{c|}{0.922}          & 0.051          \\ 
Blur                   & \multicolumn{1}{c|}{0.495}          & \multicolumn{1}{c|}{0.849}          & \multicolumn{1}{c|}{\textbf{0.001}} & \multicolumn{1}{c|}{0.473}          & \multicolumn{1}{c|}{0.892}          & 0.006          \\ 
Grayscale               & \multicolumn{1}{c|}{\textbf{0.961}} & \multicolumn{1}{c|}{0.935}          & \multicolumn{1}{c|}{0.055}          & \multicolumn{1}{c|}{\textbf{0.975}} & \multicolumn{1}{c|}{0.960}          & 0.047          \\ 
Solarization                & \multicolumn{1}{c|}{0.824}          & \multicolumn{1}{c|}{0.907}          & \multicolumn{1}{c|}{0.014}          & \multicolumn{1}{c|}{0.822}          & \multicolumn{1}{c|}{0.925}          & 0.023          \\ 
Rotation@30            & \multicolumn{1}{c|}{0.746}          & \multicolumn{1}{c|}{0.759}          & \multicolumn{1}{c|}{0.007}          & \multicolumn{1}{c|}{0.730}          & \multicolumn{1}{c|}{0.773}          & 0.012          \\ 
Rotation@60            & \multicolumn{1}{c|}{0.670}          & \multicolumn{1}{c|}{\textbf{0.744}} & \multicolumn{1}{c|}{0.002}          & \multicolumn{1}{c|}{0.643}          & \multicolumn{1}{c|}{\textbf{0.761}} & \textbf{0.004} \\ 
Rotation@90           & \multicolumn{1}{c|}{0.785}          & \multicolumn{1}{c|}{0.898}          & \multicolumn{1}{c|}{0.005}          & \multicolumn{1}{c|}{0.746}          & \multicolumn{1}{c|}{0.904}          & 0.017          \\ 
Rotation@180           & \multicolumn{1}{c|}{0.869}          & \multicolumn{1}{c|}{0.925}          & \multicolumn{1}{c|}{0.011}          & \multicolumn{1}{c|}{0.812}          & \multicolumn{1}{c|}{0.926}          & 0.027          \\ 
Horizontal Flipping                   & \multicolumn{1}{c|}{0.955}          & \multicolumn{1}{c|}{0.948}          & \multicolumn{1}{c|}{0.028}          & \multicolumn{1}{c|}{0.960}          & \multicolumn{1}{c|}{0.955}          & 0.042          \\ \hline
                        & \multicolumn{6}{c|}{\textit{SimCLRv2 (2X)}}                                                                                                                                                                  \\ \hline
Color Jitter           & \multicolumn{1}{c|}{0.940}          & \multicolumn{1}{c|}{0.865}          & \multicolumn{1}{c|}{0.086}          & \multicolumn{1}{c|}{0.960}          & \multicolumn{1}{c|}{0.883}          & 0.077          \\ 
Blurring                    & \multicolumn{1}{c|}{0.510}          & \multicolumn{1}{c|}{0.774}          & \multicolumn{1}{c|}{\textbf{0.009}} & \multicolumn{1}{c|}{0.470}          & \multicolumn{1}{c|}{0.827}          & \textbf{0.021} \\ 
Grayscale               & \multicolumn{1}{c|}{\textbf{0.948}} & \multicolumn{1}{c|}{0.887}          & \multicolumn{1}{c|}{0.111}          & \multicolumn{1}{c|}{\textbf{0.973}} & \multicolumn{1}{c|}{0.928}          & 0.056          \\ 
Solarization               & \multicolumn{1}{c|}{0.807}          & \multicolumn{1}{c|}{0.862}          & \multicolumn{1}{c|}{0.044}          & \multicolumn{1}{c|}{0.814}          & \multicolumn{1}{c|}{0.880}          & 0.050          \\ 
Rotation@30            & \multicolumn{1}{c|}{0.697}          & \multicolumn{1}{c|}{0.698}          & \multicolumn{1}{c|}{0.027}          & \multicolumn{1}{c|}{0.693}          & \multicolumn{1}{c|}{0.724}          & 0.048          \\ 
Rotation@60           & \multicolumn{1}{c|}{0.605}          & \multicolumn{1}{c|}{\textbf{0.679}} & \multicolumn{1}{c|}{0.011}          & \multicolumn{1}{c|}{0.560}          & \multicolumn{1}{c|}{\textbf{0.715}} & 0.032          \\ 
Rotation@90            & \multicolumn{1}{c|}{0.731}          & \multicolumn{1}{c|}{0.827}          & \multicolumn{1}{c|}{0.016}          & \multicolumn{1}{c|}{0.668}          & \multicolumn{1}{c|}{0.841}          & 0.040          \\ 
Rotation@180           & \multicolumn{1}{c|}{0.830}          & \multicolumn{1}{c|}{0.861}          & \multicolumn{1}{c|}{0.027}          & \multicolumn{1}{c|}{0.743}          & \multicolumn{1}{c|}{0.869}          & 0.054          \\ 
Horizontal Flipping                    & \multicolumn{1}{c|}{0.944}          & \multicolumn{1}{c|}{0.908}          & \multicolumn{1}{c|}{0.044}          & \multicolumn{1}{c|}{0.948}          & \multicolumn{1}{c|}{0.913}          & 0.084          \\ \hline
\end{tabular}
}
\label{augment_ablations}
\end{table}

\begin{figure*}[t]
    \centering
    \includegraphics[width=0.9\textwidth]{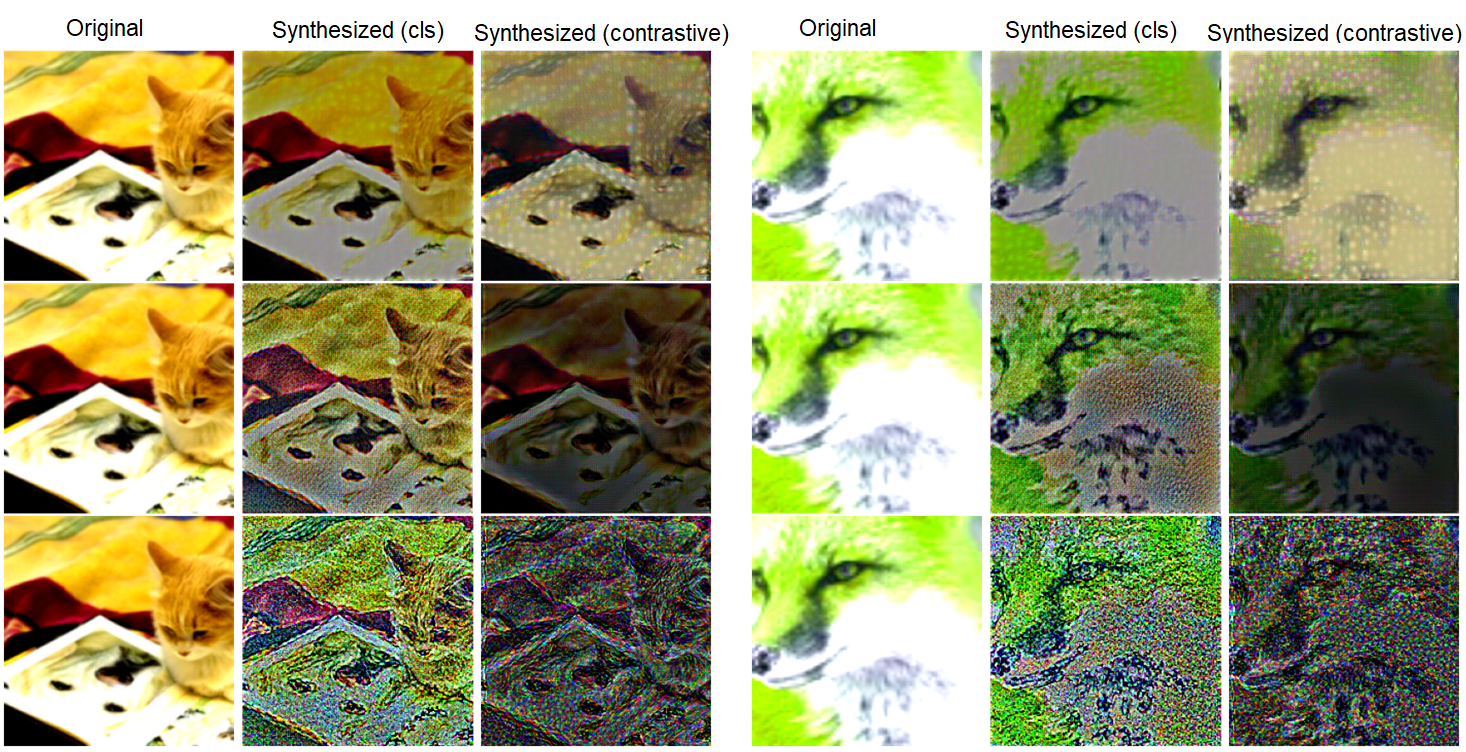}
    \caption{Contrastive models slowly learn to discard (learn to be invariant to) transformations (e.g. color) throughout the layers}
    
    \label{model_difference}
\end{figure*}

This section presents a series of observations and findings on how the contrastive model SimCLRv2~\cite{chen2020big} attributes to different visual features to infer similarity scores for pairs of images. The following findings are based on a thorough assessment of explanation maps, of which we display some representative examples for illustration. We acknowledge that these analyses are only qualitative and may omit other key points that could be discovered by reviewing a greater number of examples.

\subsubsection{Meaningless concepts and biasness towards texture} By examining some samples with our methods, we find that visual cues do not always correspond to the class of interest, which is to be expected from contrastive models. In Fig.~\ref{meaningless1}, the example shows more focus on the road than the tricycle, despite the same content being present in both images. In Fig.~\ref{texture}, tigers and cats from different image classes are easily confused due to their similar textures, which seems to suggest that the concept of striped fur is more easily learned than the animal itself. We also sampled 50 image pairs containing different objects but with a common texture (Fig.~\ref{texture}). We observe that in 64\% of the cases the methods explain similarity based on texture. 

\subsubsection{Identifying harmful data augments} We analyze a set of examples which lead to a low similarity score, with the aim of understanding the underlying cause of the low score through the use of our proposed methods. Through the analysis, we identify that the \textit{blur and blackening} augmentation harms the performance of the contrastive model and is more likely to result in bad similarity estimation. An image blackening effect is caused by a characteristic of the color jitter transform. Color jitter is characterized by alternating brightness, contrast, saturation, and hue values (in this case, alternating brightness is referred to as \textit{blackening}). Fig.~\ref{which_augment} illustrates this fact and the associated heatmaps show small overlap with any of the main subjects of the scene. That is, they do not learn any meaningful concepts. This suggests that the brightness aspect in color jitter transforms should be carefully considered. Therefore, our methods can act as tools to detect data augmentations that can be beneficial or harmful for a model. To assess the percentage of time this phenomenon happens, we sampled 50 image pairs and apply the blur and blackening effect to them. We find that 94\% of the explanations highlight meaningless concepts (Fig.~\ref{which_augment}).

\subsubsection{Dissecting Data Augmentations} Contrastive models employ a sequential approach for data augmentation, combining color jittering with grayscale and Gaussian blurring to create diverse data samples. Barlow Twins \cite{Zbontar2021BarlowTS}, on the other hand, utilizes solarization in addition to these augmentations. In this section, we aim to investigate the individual effect of each augmentation method on the model's performance, in addition to other augmentations that are not commonly used in contrastive models, such as rotations. We present an analysis in Table~\ref{augment_ablations}. By observing the top-scoring augments for each metric, we find that a 60-degree rotation for SimCLRv2 (1$\times$) has the highest impact on the model's invariance performance as measured by the explainability metrics. For SimCLRv2 (2$\times$), Gaussian blurring, grayscale transformations and 60-degree rotations are all equally impactful on the model's invariance performance. Gaussian blurring, by eliminating high-frequency details from images, and grayscale transformations, by stripping away color information, both generate more challenging examples that are meaningful for learning invariance. The results further indicate that contrastive models are also invariant to rotations, despite not being trained with this transformation.

\section{Conclusion}
\label{sec:conc}
We have proposed a set of techniques aimed at visualizing and understanding contrastive vision learning models, which operate on pairs of inputs. These techniques encompass saliency, activation, and perturbation-based approaches. For each technique, we established a baseline, which is a straightforward extension of the techniques used to explain single-image models, such as image classification models. We then highlighted the deficiencies of the baselines and proposed new explanation methods that can overcome these shortcomings. Furthermore, we have extended existing evaluation metrics to suit contrastive models. We then evaluated our techniques using these metrics and conducted an analysis, including a summary of the results, the relevance to downstream tasks, and various explorations and findings.

{\appendix[Exploring Invariance: A Visual Investigation]
In order to validate the invariance of contrastive models towards transformations like color and saturation, we conducted an experiment using feature inversion, as described in \cite{Mahendran2015UnderstandingDI}. We initialized an input image with random noise and optimized it to reconstruct features from the 3\textsuperscript{rd}, 4\textsuperscript{th}, and 5\textsuperscript{th} layers of a ResNet-50 backbone, which was trained using contrastive learning. We then compared the resulting synthesized images with those obtained from the same model trained with a different objective, namely image classification. The results are shown in Fig.~\ref{model_difference}. For the contrastive model, the image saturation diminishes with increasing layer depth, indicating that SimCLRv2 learns latent representations that are more invariant to colors resulting from the data augmentation-based training. This also aligns with the observation found in \cite{Ericsson2021HowWD} in that self-supervised models fail to retain colour information compared to supervised models. For this experiment, we use the mean squared error (MSE) as a loss function to minimize the loss of features from the synthesized image and the given feature vector at a particular layer. We initialize the synthesized image from the normal distribution with a standard deviation of 0.1 and use stochastic gradient descent (SGD) with a momentum of $0.9$ to optimize the image for $200$ iterations with a learning rate of $1e4$, which is reduced by a factor of $0.1$ every $50$ iterations. We use two different natural image priors which are derived from general properties of natural images, and which help in guiding the optimization process towards more realistic reconstructions: the total variation regularizer with a weight of $1e-8$ and the $\alpha$-norm regularizer with $\alpha$ set to $6$ and a weight of $1e-7$.}

\bibliographystyle{IEEEtran}
\bibliography{references}

 

% \end{IEEEbiography}



\vfill

\end{document}